\pdfoutput=1

\documentclass[11pt]{article}

\usepackage{acl}
\usepackage{enumitem}

\usepackage{times}
\usepackage{latexsym}

\usepackage[T1]{fontenc}

\usepackage[utf8]{inputenc}

\usepackage{microtype}

\usepackage{tabularx}
\usepackage{multirow}
\usepackage{booktabs} 
\usepackage{graphicx} 
\usepackage{float} %
\usepackage{subfigure} %
\usepackage{amsfonts}
\usepackage{textcomp,booktabs}
\usepackage{amsmath}
\usepackage{mwe}
\usepackage{colortbl}

\definecolor{maroon}{cmyk}{0,0.87,0.68,0.32}

\newcommand{\gray}{\rowcolor[gray]{.90}}

\usepackage{xcolor}

\newcommand{\ours}{\textsc{PRBoost}}

\definecolor{light-gray}{gray}{0.9}

\title{\textsc{PRBoost}: Prompt-Based Rule Discovery and Boosting\\ for Interactive Weakly-Supervised Learning}

\author{Rongzhi Zhang \\
  Georgia Tech \\
  \texttt{\small rongzhi.zhang@gatech.edu} \\\And
  Yue Yu \\
  Georgia Tech \\
  \texttt{\small yueyu@gatech.edu} \\\And
  Pranav Shetty \\
 Georgia Tech \\
  \texttt{\small pranav.shetty@gatech.edu} \\\AND
  Le Song \\
  MBZUAI \\
  \texttt{\small le.song@mbzuai.ac.ae} \\ \And  
  Chao Zhang \\
  Georgia Tech \\
  \texttt{\small chaozhang@gatech.edu} \\}

\begin{document}
\maketitle
\begin{abstract}



  Weakly-supervised learning (WSL) has shown promising results in addressing
  label scarcity on many NLP tasks, but manually designing a comprehensive,
  high-quality labeling rule set is tedious and difficult. We study
  interactive weakly-supervised learning---the problem of iteratively and
  automatically discovering novel labeling rules from data to improve the WSL
  model. Our proposed model, named \ours, achieves this goal via iterative
  prompt-based rule discovery and model boosting. It uses boosting to identify
  large-error instances and then discovers candidate rules from them by prompting
  pre-trained LMs with rule templates. The candidate rules are judged by human
  experts, and the accepted rules are used to generate complementary weak
  labels and strengthen the current model. Experiments on four tasks show
  {\ours} outperforms state-of-the-art WSL baselines up to $7.1\%$, and bridges the
  gaps with fully supervised models.Our Implementation is available at  \url{https://github.com/rz-zhang/PRBoost}.

\end{abstract}

\section{Introduction}

\emph{Weakly-supervised learning (WSL)} has recently attracted increasing
attention to mitigate the label scarcity issue in many NLP tasks. In WSL, the training data are generated by weak labeling rules obtained from sources such
as knowledge bases, frequent patterns, or human experts. The weak labeling
rules can be matched with unlabeled data to create large-scale weak labels,
allowing for training NLP models with much lower annotation cost. WSL has
recently achieved promising results in many tasks including \emph{text
  classification}
\cite{awasthi2020learning,mekala2020contextualized,meng2020text,yu2020fine},
\emph{relation extraction} \cite{zhou2020nero}, and
\emph{sequence tagging} \cite{lison2020named,safranchik2020weakly, li-etal-2021-bertifying}.

Despite its success, WSL is limited by two major factors: 1) the labeling rules, and 2) the static learning process. First, it is challenging to provide a
comprehensive and high-quality set of labeling rules a priori. Labeling rules
are often human-written \cite{ratner2017snorkel, hancock2018training}, but the
process of writing labeling rules is tedious and time-consuming even for experts.
A few works attempt to automatically discover labeling rules by mining labeled
data \cite{varma2018snuba}, or enumerating  predefined types. However, the pre-extracted rules are restricted to
frequent patterns or predefined types, which are inadequate for training an
accurate model. Second, most existing WSL methods are
static and can
suffer from the noise in the initial weak supervision~\cite{ratner2017snorkel,zhou2020nero,yu2020fine,meng2020text,zhang2022}. As the labeling rule
set remains fixed during model training, the initial errors can be
amplified, resulting in an overfitted end model. Interactive rule
  discovery has been explored in two recent works
\cite{boecking2020interactive, galhotra2021adaptive}, which solicits human
feedback on candidate rules to refine the rule set. Unfortunately, their rule
forms are limited to simple repetitive structures such as \textit{n}-grams
\cite{boecking2020interactive}, and the huge rule search space makes an
enumerating-pruning pipeline not scalable for large
datasets \cite{galhotra2021adaptive}.

Due to the above reasons, state-of-the-art WSL methods still underperform
fully-supervised methods by significant gaps on many NLP tasks. As shown in a
recent study~\cite{zhang2021wrench}, the best WSL methods fall behind the best
fully-supervised methods in 15 out of 18 NLP benchmarks; and the average performance gap is $18.84\%$ in terms of accuracy or F1 score. 


To bridge the gap between weakly-supervised and fully-supervised approaches,
we propose an iterative rule discovery and boosting framework, namely {\ours} for
interactive WSL. 
Compared to existing works on WSL and active learning, \ours~features three key designs:


First, we design a rule discovery module that uses rule templates for
prompting pre-trained language models (PLMs). By feeding difficult instances
and rule templates into PLMs, the module distills knowledge from PLMs via
prompting and generates candidate rules that capture key semantics
of the input instances. Compared to prior works based on \textit{n}-grams~\cite{boecking2020interactive}, our prompt-based rule discovery is more expressive and applicable to any tasks that support prompting.


Second, we design a boosting-style ensemble strategy to iteratively
target difficult instances and adaptively propose new rules. In each iteration,
we reweigh data by the boosting error to enforce the rule discovery
module to focus on larger-error instances. This avoids
enumerating all the possible rules and implementing post-filtering for novel rules, but
 directly targets rule discovery on large-error instances to provide
complementary information to the current model.



Third, we strategically solicit human feedback to evaluate the candidate
rules. Humans are asked to judge whether a candidate rule should be accepted
or abstained. The accepted high-quality rules are then used to generate new
weak labels that are fed into boosted model training. As the prompt-generated
rules are highly interpretable, the rule evaluation is simply a binary choice task for
human experts and thus effortless. Unlike traditional active learning methods
that annotate individual instances, such a rule-level annotation is more label-efficient because the  annotated rules can match large
amounts of instances.

We compare our method with supervised, weakly-supervised and interactive
learning baselines on four tasks: relation extraction, ontology
classification, topic classification, and chemical-protein interaction
prediction. The results show: 1) Our method outperforms state-of-the-art
weakly-supervised baselines by up to $7.1\%$; 2)
The rule-level annotation helps the model achieve higher model performance
compared to the instance-level annotation under the same budget; 3) The
machine-discovered and human-evaluated rules are of high quality, which
consistently refine the weak labels and the model in each iteration.

Our key contributions are: (1) a prompt-based rule discovery framework for
interactive WSL, which  provides flexible rule
representation while capturing subtle semantics in rule generation; (2) an
iterative boosting strategy for discovering novel rules from hard
instances and strengthening the model by an ensemble of complementary weak models;
(3) an interpretable and easy-to-annotate interactive process for rule
annotation; (4) comprehensive experiments demonstrating the
effectiveness of our framework.

\section{Related Work}
\noindent \textbf{Weakly-Supervised Learning} WSL has recently attracted much
attention in various NLP tasks. 
Despite their promising performance on various tasks, manually designing the rules can be
time-consuming. Moreover, the noise and incompleteness of the initial rules
could be propagated in model training~\cite{zhang2021wrench}. A few works attempt to reduce human
efforts in manually designing labeling rules by discovering rules from data.
For example, Snuba~\cite{varma2018snuba} generates heuristics based on a small labeled dataset with pre-defined rule types; TALLOR~\cite{li2021weakly} and
GLaRA~\cite{glara} study rule expansion for NER problem based on lexical information and then select rules based on a hand-tuned threshold. 
However, these methods discover rules in a static way and are constrained to
task-specific rule types.
In contrast, our framework discovers rules iteratively from
the entire unlabeled dataset, which can refine the rule set and enlarge its
diversity on-the-fly.




\noindent \textbf{Interactive Learning} 
Our work is related to \emph{active learning} (AL) as both involve human
annotators in the learning process. 
However, the key difference 
is that AL labels \textit{instances} based on various
query policies~\cite{holub2008entropy,shen-etal-2017-deep,zhang2020seqmix,dor2020active,margatina2021active,yu2021atm}, while our method does not
annotate individual instances, but uses annotated
rules to match unlabeled data. This makes our method more label-efficient in
leveraging human feedback for creating large-scale labeled data. 
To the best of our knowledge, only a few works have studied \emph{interactive
WSL} \cite{boecking2020interactive,galhotra2021adaptive,choi2021tagruler,hsieh2022nemo} as in our problem. However, they either use simple n-gram based rules
\cite{boecking2020interactive,hsieh2022nemo} that fail to capture sentence-level
semantics, or suffer from a huge searching space for context-free grammar
rules \cite{galhotra2021adaptive}. Unlike these works, our method uses
flexible rule representations based on prompts, and also uses boosting for targeted 
rule discovery to avoid enumerating all possible rules and performing post-filtering for novel rules.



\noindent \textbf{Language Model Prompting} Our work is also related to prompt-based learning for PLMs,
which converts the original task to a cloze-style task and leverages PLMs to fill the missing information~\cite{brown2020language,liu2021pre}. 
Prompting has been explored in various tasks,
including text classification~\cite{hu2021knowledgeable,han2021ptr,schick-schutze-2021-exploiting,schick2021just},
information
extraction~\cite{lester2021power,chen2021adaprompt} and text
generation~\cite{do2021gsum,li2021prefix}. Recent works focus on
generating better prompt templates or learning implicit prompt embeddings \cite{gao2021making, liu2021p,liu2021gpt}.
However, none of these works studied prompting for  generating weak labels.
Our work is orthogonal to them since we do not aim to optimize prompts for the original task, but uses prompts and PLMs as a knowledge source for rule discovery.



\section{Preliminaries}
\textbf{Problem Formulation }
Weakly-supervised learning (WSL) creates weak labels for model training by
applying labeling rules over unlabeled instances $D_u$. Given an unlabeled
instance $\boldsymbol{x} \in D_u$, a labeling rule $r(\cdot)$ maps $\boldsymbol{x}$ into an extended
label space: $r(\boldsymbol{x}) \rightarrow y \in \mathcal{Y} \cup\{0\}$. Here
$\mathcal{Y}$ is the original label set for the task, and $0$ is a
special label indicating $\boldsymbol{x}$ is unmatchable by $r$. Given a set
$\mathcal{R}$ of labeling rules, we can apply each rule in $\mathcal{R}$ on unlabeled instances to create a weakly labeled dataset
$\mathcal{D}_l^{\prime}$.

However, the initial weak labels $\mathcal{D}_l^{\prime}$
can be highly noisy and incomplete, which hinder the performance of WSL. We thus study the problem of \textit{interactive}
WSL: \emph{how can we automatically discover more high-quality labeling
rules to enhance the performance of WSL?} 
Besides $D_u$ and $\mathcal{D}_l^{\prime}$, we also assume access to a small
set of clean labels $\mathcal{D}_l$ $(|\mathcal{D}_l| \ll |\mathcal{D}_u|)$,
and the task is to iteratively find a set of new rules for
model improvement. In each iteration $t$, we assume a fixed rule annotation
budget $\mathcal{B}$, \textit{i.e.}, one can propose at most $\mathcal{B}$
candidate rules $\mathcal{R}_t = \{r_j\}_{j=1}^{\mathcal{B}}$ to human experts
for deciding whether each rule should
be accepted or not. The accepted rules $\mathcal{R}_t^+$ are then used to
create new weakly labeled instances $\mathcal{D}_t^{\prime}$. From
$\mathcal{D}_t^{\prime} \cup \mathcal{D}_l^{\prime}$, a model $m_t: \mathcal{X}
\rightarrow \mathcal{Y}$ can be trained to boost the performance of the
current WSL model.  \\
\noindent \textbf{Rule Representation} 
Multiple rule representations have been proposed in WSL for NLP tasks. For
example, \emph{keyword-based rules} are widely used to map
certain keywords to their highly correlated labels \cite{boecking2020interactive, meng2020text, mekala2020contextualized,bond}. \emph{Regular expression}
is another common rule format, which matches instances with pre-defined
surface patterns \cite{awasthi2020learning, yu2020fine,zhou2020nero}. \emph{Logical
  rules}~\cite{hu2016harnessing,li2021weakly} perform logical
operations (such as conjunction $\wedge$ and negation $\neg$) over atomic
rules and can thus capture higher-order compositional patterns.

We adopt a prompt-based rule representation
(Section~\ref{sec:prompt_based_rule_proposal}), which is flexible to encompass
any existing rule representations. Our prompt-based rule relies on a rule template $\tau(\cdot)$ for the target task, which contains a \texttt{[MASK]} token to be filled by a PLM $\mathcal{M}$ along with an unlabeled instance $\boldsymbol{x}$. From the
rule template $\tau$ , each candidate rule can be
automatically derived by $r = g(\mathcal{M}, \tau, \boldsymbol{x})$. Such a prompt-based rule representation is highly flexible
and can be applied to any NLP tasks that support prompting 
(see examples in Table~\ref{tab:example_rules}).



\begin{figure*}[!htb]
  \centering
  \includegraphics[scale=0.45]{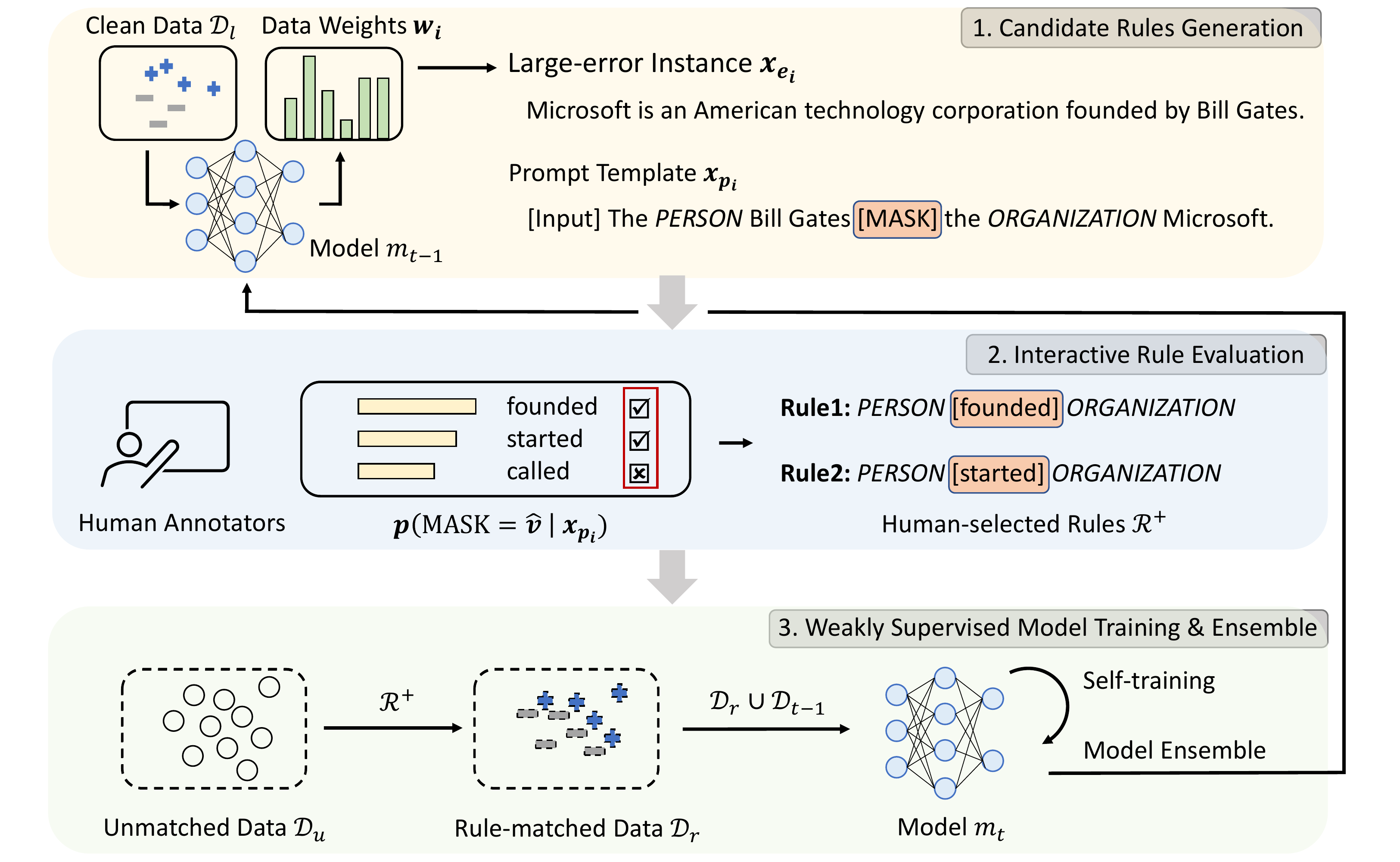}
  \vspace{-2mm}
  \caption{Overall framework for {\ours}. In each iteration, \ours~(1) identifies large-error instances from
the limited clean data and converts each large-error instance to a
prompt template for prompting-based rule discovery; (2) presents candidate
rules to human experts for annotation and uses accepted rules to generate new weak labels; (3) trains a new weak model with self-training and ensembles it with the previous models.}
  \vspace{-4mm}
  \label{fig:framework}
\end{figure*}

\section{Methodology}

\noindent\textbf{Overview}
\ours~is an iterative method for interactive WSL. In
each iteration, it proposes candidate rules from large-error
instances, solicits human feedback on candidate rules, generates weak labels,
and trains new weak models for ensembling. Figure~\ref{fig:framework}
shows the process in one iteration of \ours, which relies on three key
components:
\begin{enumerate}[leftmargin=*]
\item \emph{Candidate rule generation.} This component proposes candidate rules to be
  evaluated by human annotators. Using the small labeled dataset
  $\mathcal{D}_l$, it measures the weakness of the current model
  by identifying large-error instances on $\mathcal{D}_l$, and proposes
  rules based on these instances using PLM prompting.
\item \emph{Rule annotation and weak label creation.} This component collects human feedback to
  improve the weak supervision quality. It takes as input the candidate rules proposed by
  the previous component, and asks humans to select the high-quality
  ones. Then the human-selected rules $\mathcal{R}_t$ are used to generate
  weak labels for the unlabeled instances $\mathcal{D}_u$ in a soft-matching
  way.
\item \emph{Weakly supervised model training and ensemble}. We train a new
  weak model $m_{t+1}$ on the updated weakly labeled dataset
  $\mathcal{D}_r$. Then we self-train the weak model $m_{t+1}$ and
  integrate it into the ensemble model.
\end{enumerate}


\subsection{Candidate Rule Generation}
\label{sec:rule_generation}
\noindent\textbf{Target rule proposal on large-error instances}
We design a boosting-style \cite{hastie2009multi} strategy for generating prompt-based candidate rules. This strategy iteratively checks feature regimes in which the current model $m_t$ is weak, and proposes candidate rules from such regimes. 
We use the small labeled dataset $\mathcal{D}_l$ to identify hard instances, i.e., where the model tends to make cumulative mistakes
during iterative learning. 
The  discovered rules can complement the current rule set $\mathcal{R}$ and refine the weak labels, so the next model $m_{t+1}$ trained on the refined weakly labeled data can perform better in the weak regimes.




We initialize the weights of the instances in $\mathcal{D}_l$ as
$w_i = 1/|\mathcal{D}_l|, i = 1,2,\cdots,|\mathcal{D}_l|$.
During the
iterative model learning process,
each $w_i$ is updated as the model's weighted loss on instance
$\boldsymbol{x}_i \in \mathcal{D}_l$.
Specifically,
in iteration $t\in \{1,\cdots, n\}$, we 
weigh the samples by
\begin{equation}
   \setlength{\abovedisplayskip}{5pt}
   \setlength{\belowdisplayskip}{5pt}
  w_{i} \leftarrow w_{i} \cdot e^{\alpha_{t} \mathbb{I}\left(y_{i} \neq m_t\left(\boldsymbol{x}_i\right)\right)}, i=1,2, \ldots, |\mathcal{D}_l|. \label{eq_weights}
\end{equation}

In Equation~\ref{eq_weights}, $\alpha_t$ is the weight of model $m_t$,
which will be used for both detecting hard instances and 
model ensembling (Section~\ref{sec:ensemble}).
We compute 
$\alpha_t$
from the model's error rate on $\mathcal{D}_l$:
\begin{equation}
   \setlength{\abovedisplayskip}{1pt}
   \setlength{\belowdisplayskip}{1pt}
  \label{eq:alpha}
  \alpha_t=\log \frac{1-\operatorname{err}_t}{\operatorname{err}_t}+\log (K-1),
\end{equation}
where $\operatorname{err}_t$ is given by
\begin{equation}
   \setlength{\abovedisplayskip}{1pt}
   \setlength{\belowdisplayskip}{1pt}
  \operatorname{err}_t=\sum_{i=1}^{|\mathcal{D}_l|} w_{i} \mathbb{I}\left(y_{i} \neq m_t\left(\boldsymbol{x}_{i}\right)\right) / \sum_{i=1}^{|\mathcal{D}_l|} w_{i}.
\end{equation}


Intuitively, 
a
sample $\boldsymbol{x}_i$ 
receives a 
 larger weight $w_i$ 
 (Equation~\ref{eq_weights}) 
 if the model ensemble consistently make mistakes on $\boldsymbol{x}_i$.
A large error is often caused by poor coverage (unlabeled instances matched by
few or no rules) or dominating noise in the local feature regimes
(rule-matched labels are wrong). The weights can thus guide
the rule generator to target the top-$n$ large-error instances
$\mathcal{X}_e=\{\boldsymbol{x}_{e_i}\}_{i=1}^n$. By proposing rules from such
instances, we aim to discover novel rules 
that can
complement the current rule set and model ensemble most effectively.



\begin{table*}[]
    \centering
    \scalebox{0.925}{
\scalebox{1.0}{
\begin{tabular}{rp{15cm}}\hline
     Input :& Microsoft is an American technology corporation founded by Bill Gates. \\
     Prompt :& \texttt{[Input]} The Person
     {Bill Gates} \texttt{[Mask]} the Organization {Microsoft}.\\
     Rule :& 
     \{Entity Pair $==$ (Person, Org)\} $\land$ \{\texttt{[Mask]} $==$ \textit{founded}\} $\land$ \{$s_{t,j} \geq \text{threshold}$\} $\to$ \textbf{per:found} \\\hline
     Input :&  Marvell Software Solutions Israel is a wholly owned subsidiary of Marvell Technology Group.\\
     Prompt :& \texttt{[Input]} The {Marvell Software Solutions Israel} is a \texttt{[Mask]}.\\
     Rule :& 
     \{\texttt{[Mask]} $==$ \textit{subsidiary $\lor$ corporation $\lor$ company}\} $\land$ \{$s_{t,j} \geq \text{threshold}$\}  $\to$ \textbf{Company}\\\hline
     Input :&  Liverpool short of firepower for crucial encounter. Rafael Benitez must gamble with Liverpools Champions League prospects tonight but lacks the ammunition to make it a fair fight.\\
     Prompt :& \texttt{[Mask]} News: \texttt{[Input]} \\
     Rule :&
     \{\texttt{[Mask]} $==$ \textit{Liverpool $\lor$ Team $\lor$  Football  $\lor$  Sports}\}
     $\land$ \{$s_{t,j} \geq \text{threshold}$\} $\to$ \textbf{Sports} \\\hline
\end{tabular}
}}
  \vspace{-2mm}
  \caption{The examples of prompt-based rules for relation extraction, ontology classification, and news topic classification. Here \texttt{[Input]} denotes the original input, \texttt{[Mask]} denotes the mask token, and $\land$, $\lor$ are the logical operators. We use bold words to show the ground-truth label of the original input.}
    \label{tab:example_rules}
    \vspace{-5mm}
\end{table*}


\noindent\textbf{Prompt-based rule proposal}
\label{sec:prompt_based_rule_proposal}
For a wide range of NLP tasks such as relation extraction and text
classification, we can leverage prompts to construct informative rule
templates, which naturally leads to expressive labeling rules for WSL. 

Motivated by this, we design a rule proposal module based on PLM \emph{prompting}. 
We present concrete examples of our prompt-based rules in Table~\ref{tab:example_rules}. The input instance comes from the large-error instances identified on the clean dataset $\mathcal{D}_l$. For each task, we have a task-specific template to reshape the original input for prompting PLMs. The resulting prompt typically includes the original input as the context and a mask token to be filled by the PLMs. The final rule encompasses multiple atomic parts to capture different views of information. Each rule is accompanied by a ground-truth label of the original input instance, such a label will be assigned to the unlabeled instances matched by this rule.

For example, as shown in Table~\ref{tab:example_rules}, the prompt of the relation
extraction task can be "\emph{entity} \texttt{[MASK]} \emph{entity}", which rephrases
the original input using relation phrases while keeping the key semantics. 
Take news topic classification as another
example, by filling the masked slot in the prompt, PLMs propose candidate
keyword-based rules for topic classification. Different from the rules extracted
from surface patterns of the corpus (\textit{e.g.}, $n$-gram rules), such a prompt-based rule proposal can generate words that do not
appear in the original inputs---this capability is important to model
generalization.


Given a large-error instance $\boldsymbol{x}_{e_i} \in \mathcal{X}_e$,
we first convert it into a prompt by
$\boldsymbol{x}_{p_i} = \tau({\boldsymbol{x}_{e_i}}) \label{prompt_template}$. Such a prompt consists of the key components of the original input and a \texttt{[MASK]} token. By inheriting the original input, we construct context for the \texttt{[MASK]} token to be predicted by a pre-trained LM $\mathcal{M}$.
To complete the rule, we feed each $\boldsymbol{x}_{p_i}$ to $\mathcal{M}$ to obtain the probability distribution of the \texttt{[MASK]} token over the vocabulary $\mathcal{V}$:
\begin{align}
 \setlength{\abovedisplayskip}{1pt}
\setlength{\belowdisplayskip}{1pt}
\begin{small}
  \label{prob_task}
  p(\texttt{MASK} = \hat{\mathbf{v}}\mid\boldsymbol{x}_{p_i})
  = \frac{\exp \left(\hat{\mathbf{v}} \cdot \mathcal{M}(\boldsymbol{x}_{p_i})\right)}{\sum\limits_{\mathbf{v} \in \mathcal{V}} \exp \left(\mathbf{v} \cdot \mathcal{M}(\boldsymbol{x}_{p_i}\right))},
 \end{small}
\end{align}
where $\mathcal{M}(\cdot)$ denotes the output vector of $\mathcal{M}$, $\mathbf{v}$ is the embedding of the token in the vocabulary $\mathcal{V}$, and $\hat{\mathbf{v}}$ is the embedding of the predicted masked token.  We collect the top-$k$ predictions with highest $p(\texttt{MASK} = \hat{\mathbf{v}} \mid \boldsymbol{x}_{p_i})$
to form the candidate rules. By filling the rules based on $\boldsymbol{x}_{e_i}$ with the prompt predictions, we obtain the candidate rule set in iteration $t$, denoted as $\mathcal{R}_t = \{r_{j}\}_{j=1}^{\mathcal{B}}$.
\subsection{Rule Annotation and Matching}
\noindent\textbf{Interactive rule evaluation}
As the candidate rules $\mathcal{R}_t$ can be still noisy, \ours~thus presents
$\mathcal{R}_t$ to humans for selecting high-quality rules. Specifically, for each candidate rule $r_j\in R_t$, we present it along with its prompt template
$\boldsymbol{x}_{p_j}$ to human experts, then they judge whether the rule $r_j$ should be accepted or not. Formally, $r_j$ is associated with a label $d_j \in \{1, 0\}$. When a rule
is accepted ($d_j=1$), it will be incorporated into the accepted rule set
$\mathcal{R}^+$ for later weak label generation.






\noindent\textbf{Weak Label Generation} After human evaluation, the accepted
rules $\mathcal{R}_t^{+}$ are used to match unlabeled instances $\mathcal{D}_u$. We design
a mixed soft-matching procedure for matching rules with unlabeled instances,
which combines embedding-based similarity and prompt-based vocabulary
similarity. The two similarities complements each other: the
embedding-based similarity captures global semantics, while the prompt-based
similarity captures local features in terms of vocabulary overlapping. Given a
rule $r_j\in\mathcal{R}_t^+$ and an unlabeled instance $\boldsymbol{x}_u \in \mathcal{D}_u
$, we detail the computations of the two similarities below.



First, the embedding similarity is computed as the cosine similarity between the rule and instance embeddings~\cite{zhou2020nero}:
\begin{equation}
 \setlength{\abovedisplayskip}{1pt}
\setlength{\belowdisplayskip}{1pt}
  s_{j}^a = (\mathbf{e}_u\cdot\mathbf{e}_{r_j})/
  (\|\mathbf{e}_u\|\cdot\|\mathbf{e}_{r_j}\|),
\end{equation}
where $\mathbf{e}_u$ is the instance embedding of $\boldsymbol{x}_u$ and  $\mathbf{e}_{r_j}$ is the rule embedding of $r_j$, both embeddings are obtained from a PLM encoder.




Next, to compute the prompt-based similarity, we feed
$\tau(\boldsymbol{x}_u)$ into the prompting model (Equation \ref{prob_task}) and
use the top-$k$ candidates of the \texttt{[MASK]} position as the
predicted vocabulary for instance
$\boldsymbol{x}_u$.
We measure the vocabulary overlapping between $\mathcal{V}_u$ and $\mathcal{V}_{r_j}$ as
\begin{equation}
 \setlength{\abovedisplayskip}{1pt}
\setlength{\belowdisplayskip}{1pt}
s_{j}^b = \mid \mathcal{V}_u \cap \mathcal{V}_{r_j}\mid/k,
\end{equation}
where $\mathcal{V}_u$ is the vocabulary of instance $\boldsymbol{x}_u$ and $\mathcal{V}_{r_j}$ is the vocabulary of rule $r_j$. Note that for the unlabeled instance, we have $|\mathcal{V}_u| = k$, while for the rule, we have $|\mathcal{V}_{r_j}| \leq k$ because human annotators may abstain some candidate predictions.

The final \emph{matching score} is computed by combining the above two similarities:
 \begin{equation}
 \setlength{\abovedisplayskip}{1pt}
\setlength{\belowdisplayskip}{1pt}
  s_{j} = \alpha s_{j}^a + (1-\alpha)s_{j}^b.
\end{equation}

The instance $\boldsymbol{x}_u$ is matched by the rule $r_j$ if $s_{j}$ is higher than the matching threshold $\sigma$ obtained on the development set.
When $\boldsymbol{x}_u$ is matched by
multiple rules that provide conflicting labels, we use the one with the highest matching score to assign the weak label. If $\forall j \in 1,\cdots,k$, the
matching score $s_{j}$ is lower than $\sigma$, we abstain from labeling the instance
$\boldsymbol{x}_u$.


\subsection{Model Training \& Ensemble}
\label{sec:ensemble}
In iteration $t$, with the new rule-matched data $\mathcal{D}_r$, we obtain an enlarged weakly labeled dataset $\mathcal{D}_t = \mathcal{D}_{t-1} \cup \mathcal{D}_r$. We fit a weak model $m_t$ on $\mathcal{D}_t$ by optimizing:
\begin{equation}
 \setlength{\abovedisplayskip}{0.1pt}
\setlength{\belowdisplayskip}{0.1pt}
  \min _{\theta} \frac{1}{\left|\mathcal{D}_{t}\right|} \sum\limits_{\left(\boldsymbol{x}_i,\hat{y}_{i}\right) \in\mathcal{D}_{t}} \ell_{\operatorname{CE}}\left(m_t(\boldsymbol{x}_i), \hat{y}_{i}\right),
\end{equation}
where $\hat{y}_i$ is the weak label for instance $\boldsymbol{x}_i$,  and $\ell_{\operatorname{CE}}$ is the cross entropy loss.

While the weakly labeled dataset has been enlarged, there are still unmatched
instances in $D_u$. To exploit such unlabeled and unmatched instances, we
adopt the self-training technique for weak model training~\cite{lee2013pseudo}. The
self-training process can propagate information from the matched weak labels to the unmatched instances to improve the model
$m_t$. Following previous models \cite{xie2016unsupervised,yu2020fine}, 
for each instance $\boldsymbol{x}_i\in\mathcal{D}_u$, we generate a soft pseudo-label $\widetilde{\boldsymbol{y}}_{ij}$  from the current model $m_t$:
\begin{equation}
\setlength{\abovedisplayskip}{2pt}
\setlength{\belowdisplayskip}{2pt}
    \widetilde{\boldsymbol{y}}_{ij}=\frac{q_{ij}^2/f_j}{\sum_{j^{\prime} \in \mathcal{Y}}(q_{ij^{\prime}}^2/f_{j^{\prime}})},\quad f_j = \sum_iq_{ij}
\end{equation}
where $q_{i} = m_t(\boldsymbol{x}_i)$ is a probability vector such that $q_{i}\in \mathbb{R}^{K}$, and $q_{ij}$ is the $j$-th entry, $j\in1,\cdots,K$.

The above process yields a pseudo-labeled $\widetilde{\mathcal{D}}_u$. We update $m_t$ by optimizing:
\begin{equation}
 \setlength{\abovedisplayskip}{2pt}
\setlength{\belowdisplayskip}{2pt}
  \mathcal{L}_{c}(m_t, \widetilde{\boldsymbol{y}})=\frac{1}{|\widetilde{\mathcal{D}}_u|} \sum\limits_{\boldsymbol{x} \in  \widetilde{\mathcal{D}}_u} \mathcal{D}_{\mathrm{KL}}(\widetilde{\boldsymbol{y}} \| m_t(\boldsymbol{x})),
\end{equation}
where $\mathcal{D}_{\mathrm{KL}}(P \| Q)=\sum_{k} p_{k} \log ({p_{k}}/{q_{k}})$ is the Kullback-Leibler divergence.

Finally, we incorporate the self-trained weak model into the ensemble model. The final model is a weighted ensemble of the weak models:
\begin{equation}
 \setlength{\abovedisplayskip}{0.15pt}
\setlength{\belowdisplayskip}{0.15pt}
  f_{\theta}(\cdot) = \sum_t^n \alpha_t m_t,
  \label{eq:ensemble}
\end{equation}
where a weak model $m_t$ with a low error rate $\operatorname{err}_t$ will be assigned a higher coefficient $\alpha_t$ according to Equation~\ref{eq:alpha}.
\section{Experiments}
\label{sec:results}

\begin{table*}[!htb]
    \centering
    \scalebox{0.85}{
\begin{tabular}{lcccc}\toprule
    \textbf{Method (Metrics)} & \textbf{TACRED (F1)} & \textbf{DBpedia (Acc.)} & \textbf{ChemProt (Acc.)} & \textbf{AG News (Acc.)}  \\\midrule
    $\textbf{Supervised Baselines}$&&&\\
    PLM w. 100\% training data&66.9 (66.3/67.6) &99.4 &79.7 & 94.4\\
    PLM w. limited training data$^\dagger$ & 32.9 (40.8/27.6) & 98.0 & 59.4 &86.4\\\midrule
    \textbf{Weakly Supervised Baselines} & & & &\\
    Rule Matching & 20.1 (\textbf{85.0}/11.4) & 63.2 & 46.9 & 52.3 \\
    Snorkel \cite{ratner2017snorkel}& \textcolor{blue}{39.7} (39.2/40.1) & 69.5 & 56.4 & 86.2\\
    LOTClass \cite{meng2020text}& --- &\textcolor{blue}{91.1} & --- &86.4\\
    COSINE \cite{yu2020fine}& 39.5 (38.9/40.3) & 73.1 &\textcolor{blue}{59.8} &\textcolor{blue}{87.5}\\
        \gray
    Snorkel + fine-tuning$^\dagger$ & 40.8 (41.0/40.6) &97.6 & 64.9 & 87.7 \\
     \gray
    LOTClass + fine-tuning$^\dagger$ & --- & \textcolor{purple}{98.1} & --- & \textcolor{purple}{88.0}\\
    \gray
    COSINE + fine-tuning$^\dagger$ &  \textcolor{purple}{41.0} (40.4/41.7) & 97.9 & \textcolor{purple}{65.7} & \textcolor{purple}{88.0} \\\midrule
    \textsc{PRBoost} &\textbf{48.1} (42.7/\textbf{55.1)} & \textbf{98.3} & \textbf{67.1} &\textbf{88.9}\\\bottomrule
\end{tabular}
}

    \caption{Main results on four benchmark datasets. $\dagger$: we use different proportions of clean data for fine-tuning as described in Section~\ref{exp_setup}. We use gray background to show the results of WLS baselines fine-tuned on the clean data. We highlight the best fine-tuned results with purple font, and the best WSL results with blue font.}
    \vspace{-4mm}
    \label{tab:main_res}
\end{table*}

\begin{figure*}[t]
    \vspace{-2mm}
        \centering
        \subfigure[Iteration $0$]{
            \includegraphics[width=0.245\textwidth]{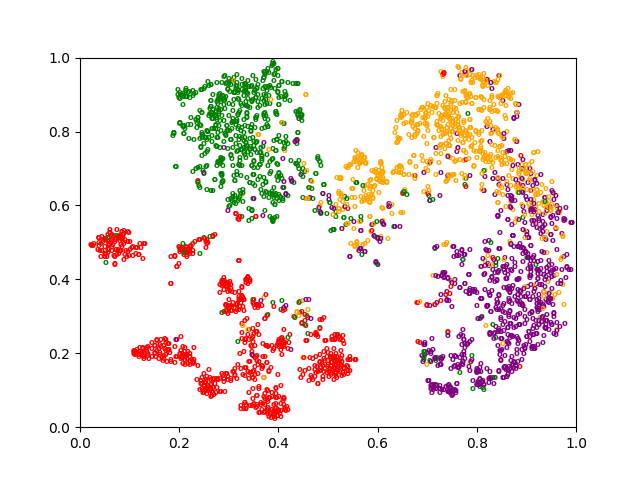}
            \label{fig:vis_sub0}
        }\hspace{-3mm}
        \subfigure[Iteration $1$]{
            \includegraphics[width=0.245\textwidth]{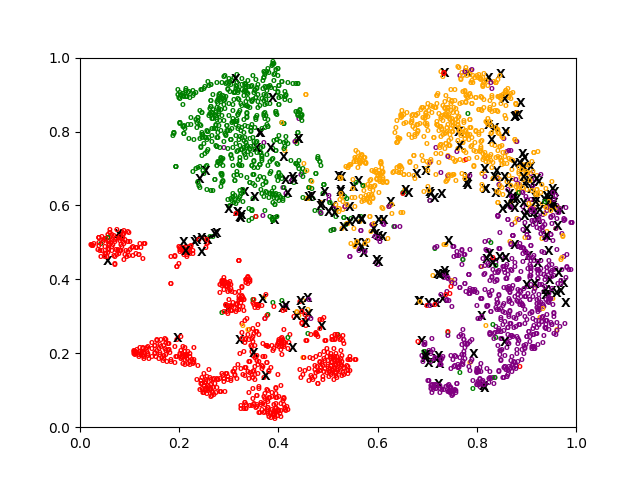}
            \label{fig:vis_sub1}
        }\hspace{-3mm}
        \subfigure[Iteration $4$]{
            \includegraphics[width=0.245\textwidth]{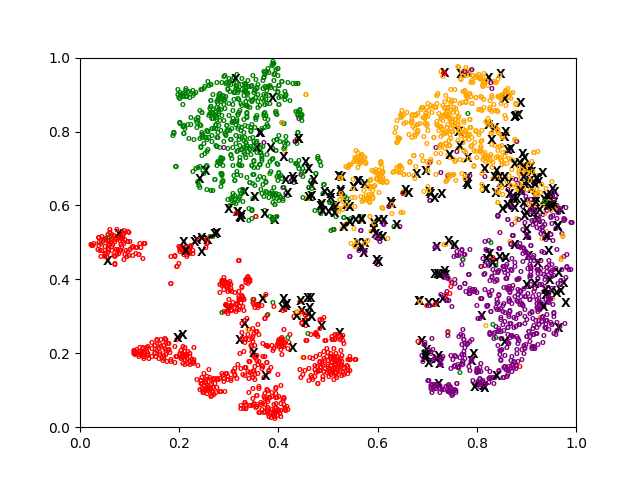}
            \label{fig:vis_sub4}
        }\hspace{-3mm}
        \subfigure[Iteration $10$]{
            \includegraphics[width=0.245\textwidth]{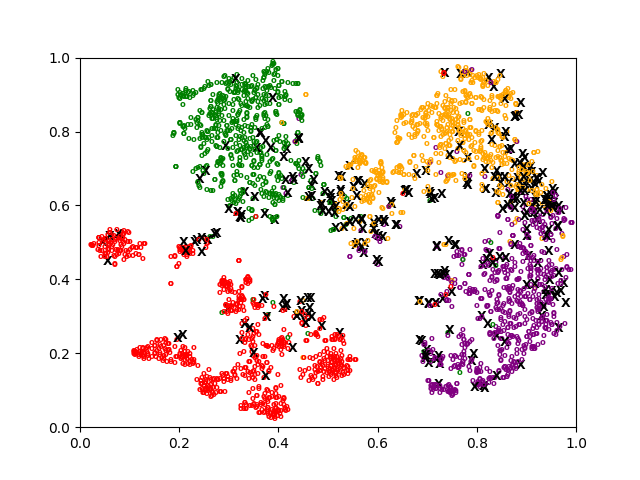}
            \label{fig:vis_sub10}
        }
            \vspace{-2ex}
        \caption{T-SNE visualization \cite{van2008visualizing} of rule-matched data that mis-classified by the model on AG News dataset. The four classes are represented by different colors, and the black cross denotes the rule-matched data.  }\label{fig:vis_rule}
         \vspace{-3ex}
\end{figure*}

\subsection{Experiment Setup}
\label{exp_setup}
\noindent\textbf{Tasks and Datasets}
We conduct experiments on four benchmark datasets, including \emph{TACRED}~\cite{zhang2017position} for relation extraction, \emph{DBPedia}~\cite{zhang2015character} for ontology classification, \emph{ChemProt}~\cite{chemprot} for chemical-protein interaction classification and \emph{AG News}~\cite{zhang2015character} for news topic classification.
For the initial weak supervision sources, we use the labeling rules provided by existing works: \citet{zhou2020nero} for TACRED, \citet{meng2020text} for DBPedia, and \citet{zhang2021wrench} for Chemprot and AG News.
The statistics of the four datasets are shown in table~\ref{tab:dataset}. For the development set, we do not directly use the full development set as suggested by the recent works \citep{gao2021making,perez2021true}. This prevents the model from taking the advantage of the massive number of labeled data in the development set. Instead, we create a real label-scarce scenario and keep the number of sample in validation set $\mathcal{D}_v$ the same as the limited clean labeled set $\mathcal{D}_l$, namely $|\mathcal{D}_v| = |\mathcal{D}_l|$.

\noindent\textbf{Baselines}
We include three groups of baselines: \\
\noindent  \textbf{Fully Supervised Baseline}: \textbf{PLM}:
  We use the pretrained language model RoBERTa-base~\cite{liu2019roberta} as the backbone and fine-tune it with the full clean  labeled data except for  ChemProt. On
ChemProt, we choose BioBERT~\cite{lee2020biobert} as the backbone for all the baselines and our model
to better adapt to this domain-specific task. 
The performance of fully supervised methods serves as an upper bound for weakly-supervised methods.

\noindent \textbf{Weakly Supervised Baselines}: (1) \textbf{Snorkel}~\cite{ratner2017snorkel} is a classic  WSL model. It aggregates different labeling functions
  with probabilistic models, then fed the  aggregated
   labels  to PLM for the target task. 
  (2) \textbf{LOTClass}~\cite{meng2020text} is a recent model for weakly-supervised text classification.  It uses label names to probe PLMs to generate weak labels, and performs self-training using the weak labels for classification.
(3) \textbf{COSINE}~\cite{yu2020fine} is a state-of-the-art method on fine-tuning
  PLMs with weak supervision.
  It adopts self-training and contrastive learning to fine-tune LMs with weakly-labeled data. 
\\ \textbf{Interactive Learning Baselines}:
\noindent (1) \textbf{Entropy}-based AL \cite{holub2008entropy} is a simple-yet-effective method for AL which acquires samples with the highest predictive entropy.
(2) \textbf{CAL}~\cite{margatina2021active} is the most recent method for active learning. It selects samples has the most diverge predictions from their neighbors for annotation.
(3) \textbf{IWS}~\cite{boecking2020interactive} is an interactive WSL model. It firstly generates n-gram terms as candidate rules, then selects quality rules by learning from humans' feedback. Note that IWS is designed for binary classification, which makes it hard to adapt to classification with multiple labels.  



\noindent\textbf{Evaluation Protocol} To propose rules on large-error
instances, we assume access to a dataset $\mathcal{D}_l$ with a limited number
of clean labeled data. For our method, such a clean dataset is only used for
identifying large-error instances. For fair comparison, for the
WSL baselines, we further fine-tune them using the same
clean data and compare with such fine-tuned results. Specifically, we use
$5\%$ clean data for TACRED and ChemProt, $0.5\%$ for AG News and $0.1\%$ for
DBPedia. We then implement a 10-iteration rule proposal and weak model
training. In each iteration, we identify the top-$10$ large-error instances and
propose 100 candidate rules in total (\textit{i.e.}, 10 candidate rules per instance). Each rule is annotated by three humans, and the annotated rule labels are majority-voted for later weak label generation. 
Following the common practice~\cite{zhang2017position,zhang2021wrench}, we use F1 score for TACRED and accuracy for other datasets.




\subsection{Main Results}

Table~\ref{tab:main_res} shows the performance of \ours~and the
baselines on the four datasets. The results show that \ours~outperforms the
weakly supervised baselines on all the four datasets. When the weakly
supervised baselines are not fine-tuned on $\mathcal{D}_l$, \ours~outperforms
the strongest WSL baseline by $8.4\%$, $7.2\%$, $7.3\%$, $2.4\%$ on the four benchmarks. Even when
the WSL models are further fine-tuned using clean labeled data, \ours~still
outperform them by $2.4\%$ on average. Compared against supervised baselines,
\textsc{PRBoost} is significantly better than the fine-tuned model on TACRED,
ChemProt and AG News when the training data is limited. For the model
fine-tuned with $100\%$ training data, we narrow the gap to fully supervised
learning, compared to other WS approaches.

Comparing the performance gains across datasets, the performance gap between
\ours~and the baselines is the largest on TACRED, which is the most
challenging task among the four with 41 different relation types. ChemProt is
the smallest dataset with only $5400$ training data, so the gain is larger
when the WSL methods are fine-tuned with clean labels. The performance gaps
among different methods are small on DBPedia, especially after they are
fine-tuned using clean labeled data. DBpedia, being  a relatively simple dataset,
using only $0.1\%$ clean data for  fine-tuning RoBERTa already achieves 98\%
accuracy, and the other WSL methods after fine-tuning perform similarly.


It is worth noting that \ours~performs strongly across all the tasks because
we can easily design a task-specific prompt template to adapt to each task. In
contrast, some WSL baselines are difficult to apply to certain tasks. For example,
LOTClass achieves strong performance for DBpedia and AGNews as its weak
sources are tailored for text classification. However, it is hard to apply it
to relation extraction tasks. Similarly, IWS performs well on binary
classification problems using n-gram based rules, but the method is only designed for binary classification, 
making it unsuitable for complex multi-class tasks.


\subsection{Rule Annotation Agreement and Cost}
\label{sec_user_exp}

\begin{figure}[]
    \centering
    \includegraphics[scale=0.36]{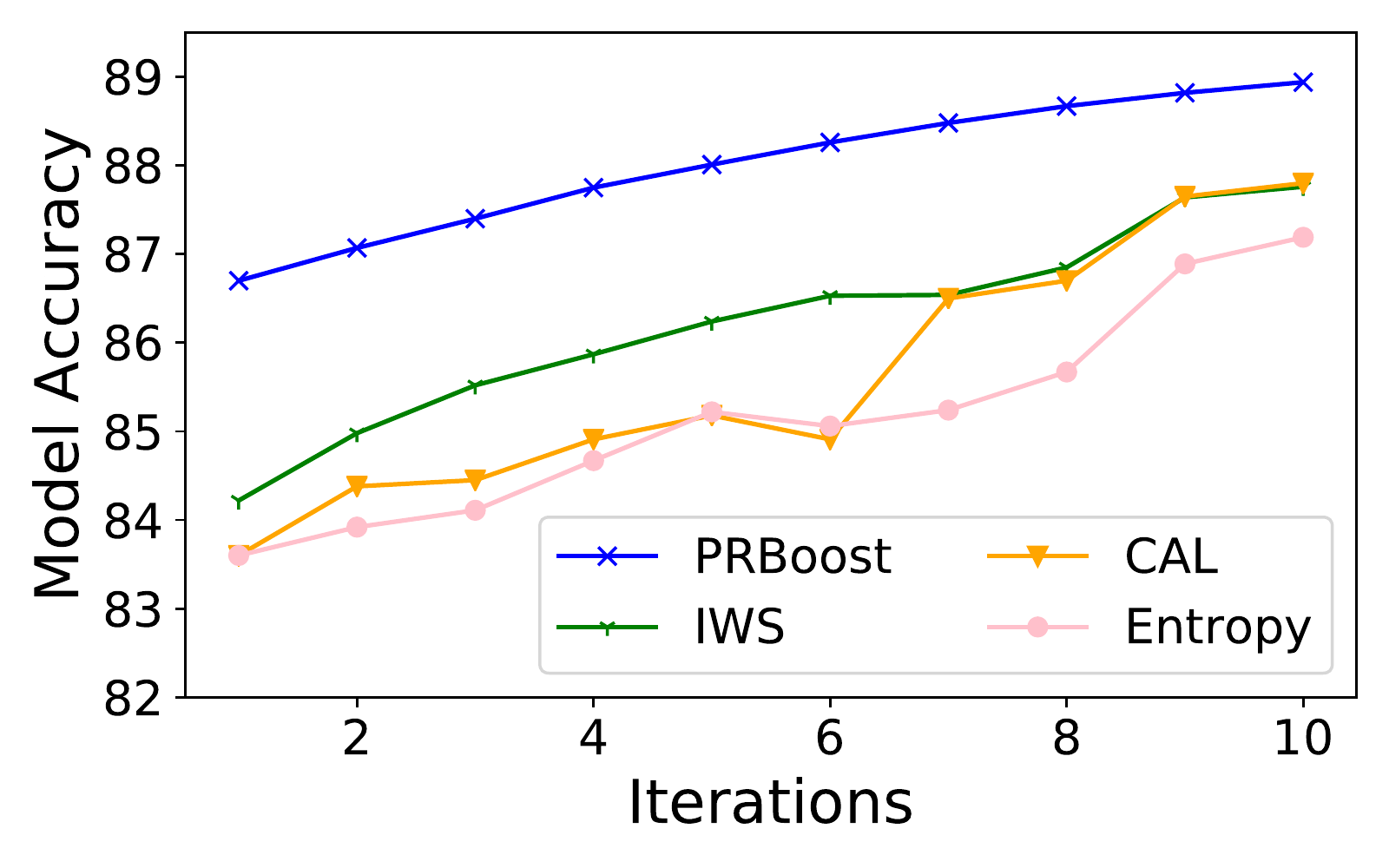}
    \vspace{-4mm}
    \caption{Results of interactive methods on AG News}
    \label{fig:annotation}
    \vspace{-6mm}
\end{figure}



In this set of experiments, we benchmark model performance and annotation cost
against interactive learning baselines (detailed in Appendix~\ref{iter_baseline}): IWS, CAL, and Entropy-based AL. As shown in
Figure~\ref{fig:annotation}, \ours~outperforms IWS that also features rule-level annotation by $1.2\%$ with very close annotation cost.
Our method outperforms the best interactive baseline CAL by $1.1\%$ in terms of accuracy, while using about $0.6\times$ annotation cost. While annotating model-proposed rule
or instances, we asked all the three annotators to time their annotation. On average, it takes each annotator less than 3 seconds to annotate
one rule, while it takes nearly 10 seconds to annotate one instance.
Rule-level annotation is much more efficient than instance-level annotation
because 1) we show the prompt rather than the original instance to humans,
which is shorter and easier to read; 2) upon scanning the prompt, the
annotators can  swiftly select qualified rules as they only
differ at the \texttt{[MASK]} position. This shows that rule-level annotation is
an efficient and suitable paradigm for interactive WSL.

\begin{table}[!htb]
    \centering
    
\scalebox{0.6}{
\begin{tabular}{c|ccccc ccccc c}\toprule
     Iteration& 1&2&3&4&5&6&7&8&9&10& Overall\\\midrule
     $\bar{P}$&.89&.90&.93&.90&.87&.92&.91&.91&.87&.90&.90\\
     $\bar{P_e}$&.63&.59&.73&.71&.62&.73&.66&.56&.68&.68&.65\\
     $\kappa$&.71&.77&.73&.66&.65&.71&.75&.79&.60&.68&.71\\\bottomrule
\end{tabular}}
    \vspace{-3mm}
    \caption{Annotation agreement measured by the Fleiss-Kappa $\kappa$ on AG News.   $\bar{P}$ measures annotation agreement over all categories; $\bar{P_e}$ computes the quadratic sum of the proportion of assignments to each category.}
    \label{tab:fleiss-kappa-metrics}
    \vspace{-3mm}
\end{table}
For the annotation agreement, we compute Fleiss' kappa $\kappa$ \cite{fleiss1971measuring} to evaluate the agreement among multiple human annotators. This statistic assesses the reliability of agreement among multiple annotators. $\kappa=1$ indicates complete agreement over all the annotators, and no agreement results in $\kappa \leq 0$.
As shown in Table~\ref{tab:fleiss-kappa-metrics}, we obtained an average $\kappa=0.71$, which means the annotators achieve substantial agreement. For each iteration, the $\kappa$ ranges between $[0.60, 0.79]$ indicating the stability of the annotation agreement.

\subsection{Rule Quality in Iterative Learning}
\label{sec_rule_qual}
\begin{figure}
    \centering
    \includegraphics[scale=0.25]{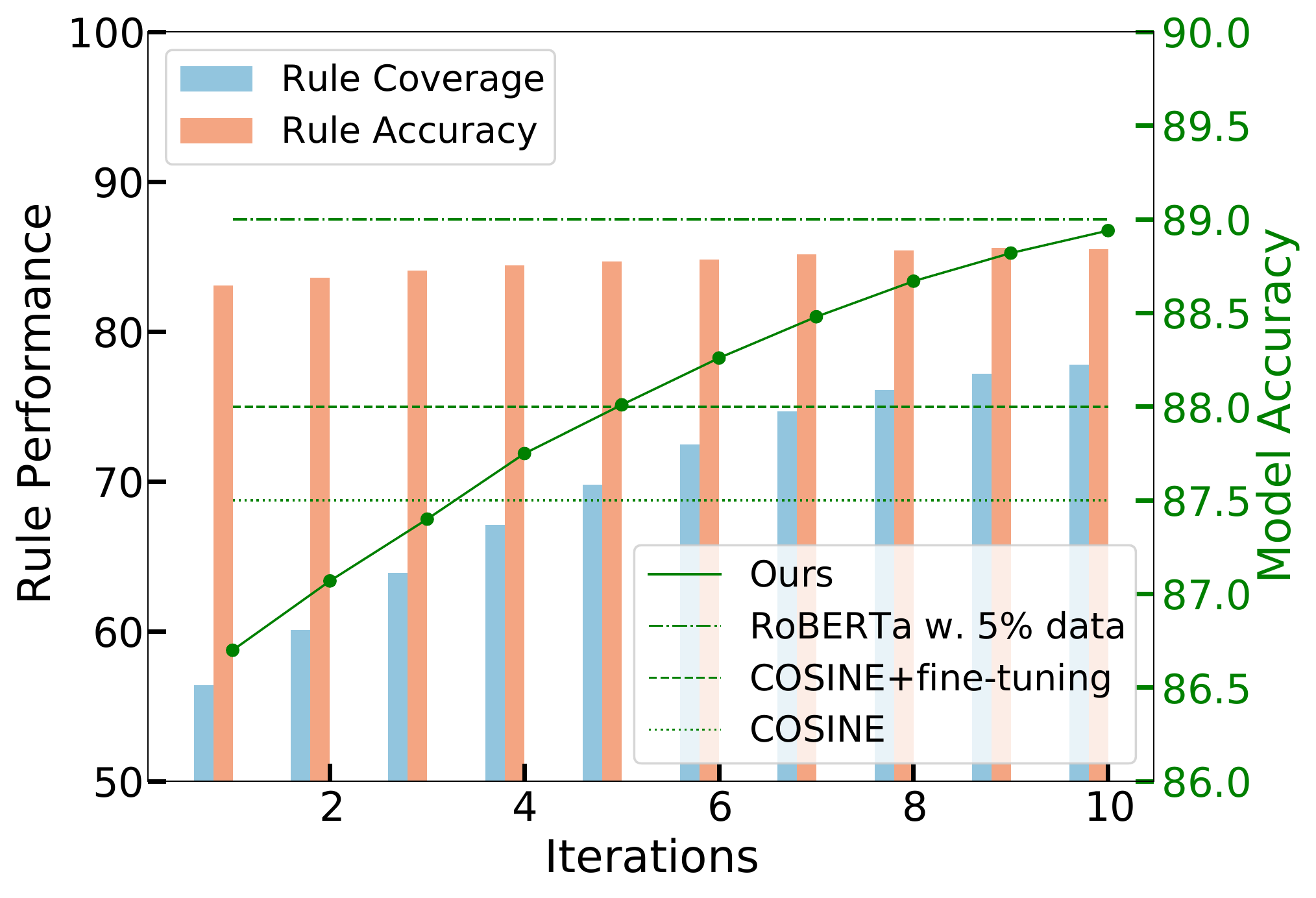}
    \vspace{-3mm}
    \caption{Rule performance and model accuracy v.s. iterations on AG News.}
    \vspace{-6mm}
    \label{fig:perf_iter}
\end{figure}

In this set of experiments, we evaluate the quality of the rules discovered by
\ours. Figure~\ref{fig:vis_rule} visualizes the discovered rules on AG News dataset.
We observe that 1) the rules can rectify some mis-classified data, and 2) the rules can complement each other.
For the first observation, we can take Figure~\ref{fig:vis_sub0} and Figure~\ref{fig:vis_sub1} for example. In iteration 0 where new rules have not been proposed, it is obvious that some green data points and purple data points are mixed into the orange cluster. After the first-round rule proposal, \ours~ has already rectified parts of wrong predictions via rule-matching. This is because our rule proposal is targeted on the large-error instances, such adaptively discovered rules can capture the model's weakness more accurately compared to the simply enumerated rules. For the second observation, we found that more mis-classified data points get matched by the newly discovered rules as the iteration increases. It demonstrates \ours~ can gradually enlarge the effective rule set by adding complementary rules, which avoids proposing repetitive rules that can not improve the rule coverage.

Figure~\ref{fig:perf_iter} shows the changes in rule accuracy, rule
coverage, and model performance in the iterative learning process on AG News.
As shown, the model's accuracy increases steadily during learning, which is
improved from $86.7\%$ to $88.9\%$ after 10 iterations. This improvement
arises from two key aspects of \ours. First, the enlarged rule set
continuously augments weakly labeled data, which provides more supervision for
the weak model training. Second, the model ensemble approach refines the
previous large errors step by step, resulting in increasing ensemble
performance.

Regarding the rule coverage and accuracy, we observe the coverage of the rule
set is improved from $56.4\%$ to $77.8\%$, and rule accuracy from $83.1\%$ to
$85.6\%$. Such improvements show that \ours~can adaptively propose novel rules
to complement the previous rule set, which can match more instances that were
previously unmatchable. Note that the increased rule converge has not
compromised rule accuracy, but rather improved it. The reason is two-fold: (1)
the human-in-the-loop evaluation can select high-quality rules for generating
new weak labels; (2) for the instances with wrong initial weak labels,
\ours~can discover more rules for the same instances and correct the weak
labels through majority voting.


\subsection{Ablation Study}
\label{sec_ablation}
We study the effectiveness of various components in \textsc{PRBoost} and show
the ablation study results in Figure~\ref{fig:ablation}. We have the
following findings:
\begin{figure}[]
    \centering
    \includegraphics[scale=0.32]{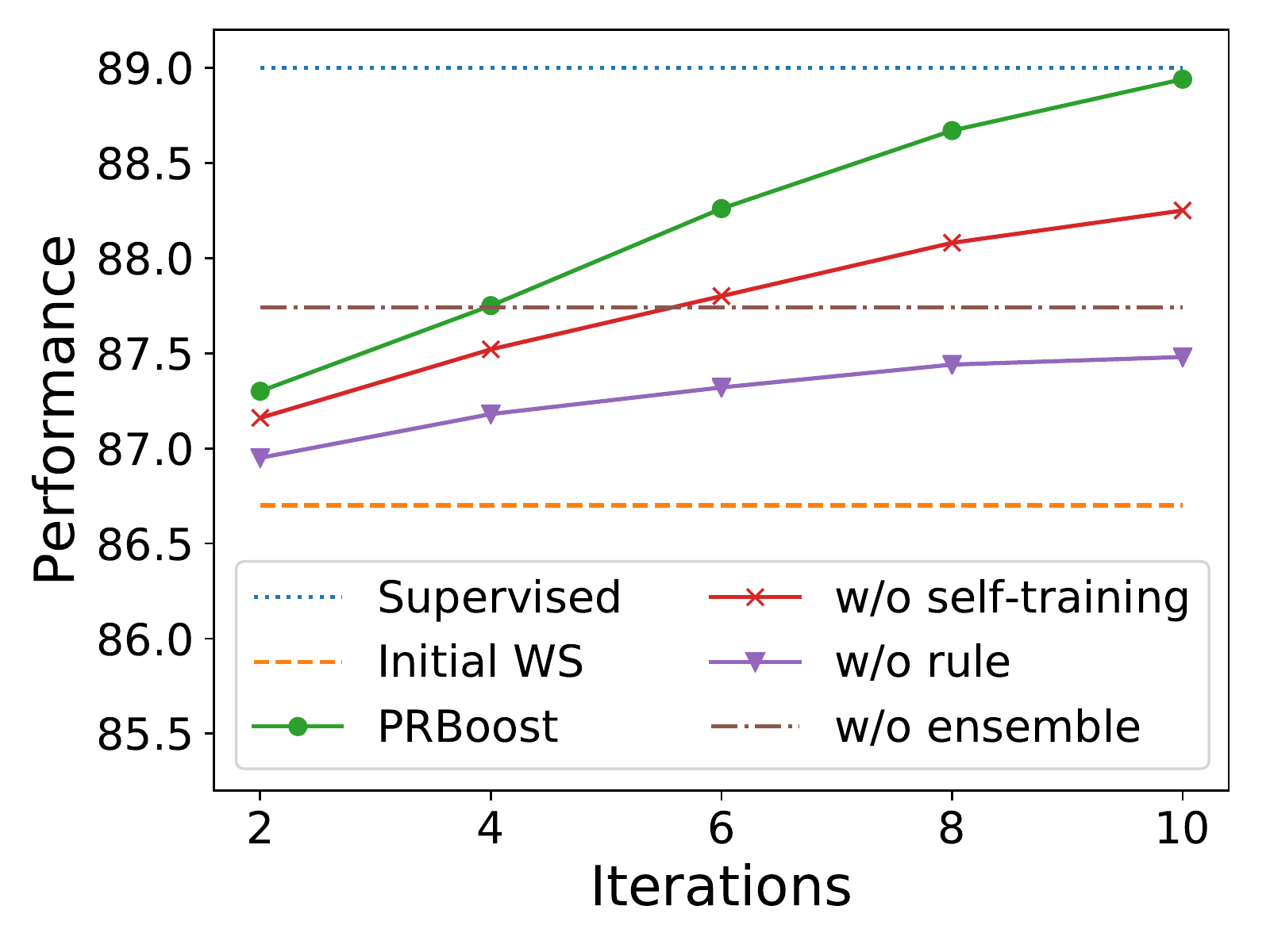}
    \vspace{-4mm}
    \caption{Ablation study on AG News. The three horizontal lines represent the no-iterative methods. We use COSINE as the initial WS baseline. For the supervised baseline, we fine-tune RoBERTa on $5\%$ clean data.}
    \vspace{-5mm}
    \label{fig:ablation}
\end{figure}

First, the boosting-based iterative rule discovery strategy is effective. For
the "w/o ensemble" setting, we fix the annotation budget $\mathcal{B}$ but
discover candidate rules from large-error samples in one iteration. The results
show the superiority of the iterative strategy in \ours~, which brings $1.2\%$
performance gain. \ours~iteratively identifies the current model's weaknesses
and proposes rules to strengthen itself, therefore it adaptively discovers
more effective rules than static rule discovery.

Second, ensembling alone without new rule discovery is not as effective. For
the "w/o rule" variant, we do not propose new rules, but ensemble multiple
self-trained weak classifiers instead. The final performance drops
significantly under this setting by $1.5\%$. It demonstrates the newly
proposed rules provide complementary weak supervision to the model. Although
simply ensembling multiple weak classifiers also helps WSL, it is not as
effective as training multiple complementary weak models as in \ours.

Third, self-training benefits learning from new weak labels. For the "w/o
self-training" setting, we do not use the self-training technique when learning
each weak classifier. The performance deteriorates by $0.6\%$. This is because
part of the data are still unmatched after we propose new rules, and
self-training leverages the unlabeled data to help the model generalize
better.

\section{Conclusion}
We proposed {\ours} to iteratively discover prompt-based rules for interactive
weakly-supervised learning. Through a boosting-style ensemble strategy, it
iteratively evaluates model weakness to identify large-error instances for new rule proposal. From such large-error instances, its prompt-based rule
discovery module leads to expressive rules that can largely improve rule
coverage while being easy to annotate. The discovered rules 
complement the current rule set and refine the WSL model continuously. Our experiments on four benchmarks demonstrate that \ours~can largely
improve WSL and narrow the gaps between WSL models and fully-supervised
models.

\section*{Acknowledgements}
This work was supported by ONR MURI N00014-17-1-2656, NSF IIS-2008334, IIS-2106961,  and research awards from Google, Amazon, Facebook, and Kolon Inc.

\bibliography{custom}

\begin{thebibliography}{53}
\expandafter\ifx\csname natexlab\endcsname\relax\def\natexlab#1{#1}\fi

\bibitem[{Awasthi et~al.(2020)Awasthi, Ghosh, Goyal, and
  Sarawagi}]{awasthi2020learning}
Abhijeet Awasthi, Sabyasachi Ghosh, Rasna Goyal, and Sunita Sarawagi. 2020.
\newblock \href {https://openreview.net/forum?id=SkeuexBtDr} {Learning from
  rules generalizing labeled exemplars}.
\newblock In \emph{International Conference on Learning Representations}.

\bibitem[{Boecking et~al.(2021)Boecking, Neiswanger, Xing, and
  Dubrawski}]{boecking2020interactive}
Benedikt Boecking, Willie Neiswanger, Eric Xing, and Artur Dubrawski. 2021.
\newblock \href {https://openreview.net/forum?id=IDFQI9OY6K} {Interactive weak
  supervision: Learning useful heuristics for data labeling}.
\newblock In \emph{International Conference on Learning Representations}.

\bibitem[{Brown et~al.(2020)Brown, Mann, Ryder, Subbiah, Kaplan, Dhariwal,
  Neelakantan, Shyam, Sastry, Askell, Agarwal, Herbert-Voss, Krueger, Henighan,
  Child, Ramesh, Ziegler, Wu, Winter, Hesse, Chen, Sigler, Litwin, Gray, Chess,
  Clark, Berner, McCandlish, Radford, Sutskever, and
  Amodei}]{brown2020language}
Tom Brown, Benjamin Mann, Nick Ryder, Melanie Subbiah, Jared~D Kaplan, Prafulla
  Dhariwal, Arvind Neelakantan, Pranav Shyam, Girish Sastry, Amanda Askell,
  Sandhini Agarwal, Ariel Herbert-Voss, Gretchen Krueger, Tom Henighan, Rewon
  Child, Aditya Ramesh, Daniel Ziegler, Jeffrey Wu, Clemens Winter, Chris
  Hesse, Mark Chen, Eric Sigler, Mateusz Litwin, Scott Gray, Benjamin Chess,
  Jack Clark, Christopher Berner, Sam McCandlish, Alec Radford, Ilya Sutskever,
  and Dario Amodei. 2020.
\newblock \href
  {https://proceedings.neurips.cc/paper/2020/file/1457c0d6bfcb4967418bfb8ac142f64a-Paper.pdf}
  {Language models are few-shot learners}.
\newblock In \emph{Advances in Neural Information Processing Systems},
  volume~33, pages 1877--1901.

\bibitem[{Chen et~al.(2021)Chen, Xie, Zhang, Yan, Deng, Tan, Huang, Si, and
  Chen}]{chen2021adaprompt}
Xiang Chen, Xin Xie, Ningyu Zhang, Jiahuan Yan, Shumin Deng, Chuanqi Tan, Fei
  Huang, Luo Si, and Huajun Chen. 2021.
\newblock Adaprompt: Adaptive prompt-based finetuning for relation extraction.
\newblock \emph{arXiv preprint arXiv:2104.07650}.

\bibitem[{Choi et~al.(2021)Choi, Evensen, Demiralp, and
  Hruschka}]{choi2021tagruler}
Dongjin Choi, Sara Evensen, {\c{C}}a{\u{g}}atay Demiralp, and Estevam Hruschka.
  2021.
\newblock Tagruler: Interactive tool for span-level data programming by
  demonstration.
\newblock In \emph{Companion Proceedings of the Web Conference 2021}, pages
  673--677.

\bibitem[{Dou et~al.(2021)Dou, Liu, Hayashi, Jiang, and Neubig}]{do2021gsum}
Zi-Yi Dou, Pengfei Liu, Hiroaki Hayashi, Zhengbao Jiang, and Graham Neubig.
  2021.
\newblock \href {https://doi.org/10.18653/v1/2021.naacl-main.384} {{GS}um: A
  general framework for guided neural abstractive summarization}.
\newblock In \emph{Proceedings of the 2021 Conference of the North American
  Chapter of the Association for Computational Linguistics: Human Language
  Technologies}, pages 4830--4842, Online. Association for Computational
  Linguistics.

\bibitem[{Ein-Dor et~al.(2020)Ein-Dor, Halfon, Gera, Shnarch, Dankin, Choshen,
  Danilevsky, Aharonov, Katz, and Slonim}]{dor2020active}
Liat Ein-Dor, Alon Halfon, Ariel Gera, Eyal Shnarch, Lena Dankin, Leshem
  Choshen, Marina Danilevsky, Ranit Aharonov, Yoav Katz, and Noam Slonim. 2020.
\newblock \href {https://doi.org/10.18653/v1/2020.emnlp-main.638} {{A}ctive
  {L}earning for {BERT}: {A}n {E}mpirical {S}tudy}.
\newblock In \emph{Proceedings of the 2020 Conference on Empirical Methods in
  Natural Language Processing (EMNLP)}, pages 7949--7962, Online. Association
  for Computational Linguistics.

\bibitem[{Fleiss(1971)}]{fleiss1971measuring}
Joseph~L Fleiss. 1971.
\newblock Measuring nominal scale agreement among many raters.
\newblock \emph{Psychological bulletin}, 76(5):378.

\bibitem[{Galhotra et~al.(2021)Galhotra, Golshan, and
  Tan}]{galhotra2021adaptive}
Sainyam Galhotra, Behzad Golshan, and Wang-Chiew Tan. 2021.
\newblock Adaptive rule discovery for labeling text data.
\newblock In \emph{Proceedings of the 2021 International Conference on
  Management of Data}, pages 2217--2225.

\bibitem[{Gao et~al.(2021)Gao, Fisch, and Chen}]{gao2021making}
Tianyu Gao, Adam Fisch, and Danqi Chen. 2021.
\newblock \href {https://doi.org/10.18653/v1/2021.acl-long.295} {Making
  pre-trained language models better few-shot learners}.
\newblock In \emph{Proceedings of the 59th Annual Meeting of the Association
  for Computational Linguistics and the 11th International Joint Conference on
  Natural Language Processing (Volume 1: Long Papers)}, pages 3816--3830.
  Association for Computational Linguistics.

\bibitem[{Han et~al.(2021)Han, Zhao, Ding, Liu, and Sun}]{han2021ptr}
Xu~Han, Weilin Zhao, Ning Ding, Zhiyuan Liu, and Maosong Sun. 2021.
\newblock Ptr: Prompt tuning with rules for text classification.
\newblock \emph{arXiv preprint arXiv:2105.11259}.

\bibitem[{Hancock et~al.(2018)Hancock, Varma, Wang, Bringmann, Liang, and
  R{\'e}}]{hancock2018training}
Braden Hancock, Paroma Varma, Stephanie Wang, Martin Bringmann, Percy Liang,
  and Christopher R{\'e}. 2018.
\newblock \href {https://doi.org/10.18653/v1/P18-1175} {Training classifiers
  with natural language explanations}.
\newblock In \emph{Proceedings of the 56th Annual Meeting of the Association
  for Computational Linguistics (Volume 1: Long Papers)}, pages 1884--1895,
  Melbourne, Australia. Association for Computational Linguistics.

\bibitem[{Hastie et~al.(2009)Hastie, Rosset, Zhu, and Zou}]{hastie2009multi}
Trevor Hastie, Saharon Rosset, Ji~Zhu, and Hui Zou. 2009.
\newblock Multi-class adaboost.
\newblock \emph{Statistics and its Interface}, 2(3):349--360.

\bibitem[{Holub et~al.(2008)Holub, Perona, and Burl}]{holub2008entropy}
Alex Holub, Pietro Perona, and Michael~C Burl. 2008.
\newblock Entropy-based active learning for object recognition.
\newblock In \emph{2008 IEEE Computer Society Conference on Computer Vision and
  Pattern Recognition Workshops}, pages 1--8. IEEE.

\bibitem[{Hsieh et~al.(2022)Hsieh, Zhang, and Ratner}]{hsieh2022nemo}
Cheng-Yu Hsieh, Jieyu Zhang, and Alexander Ratner. 2022.
\newblock Nemo: Guiding and contextualizing weak supervision for interactive
  data programming.
\newblock \emph{arXiv preprint arXiv:2203.01382}.

\bibitem[{Hu et~al.(2021)Hu, Ding, Wang, Liu, Li, and
  Sun}]{hu2021knowledgeable}
Shengding Hu, Ning Ding, Huadong Wang, Zhiyuan Liu, Juanzi Li, and Maosong Sun.
  2021.
\newblock Knowledgeable prompt-tuning: Incorporating knowledge into prompt
  verbalizer for text classification.
\newblock \emph{arXiv preprint arXiv:2108.02035}.

\bibitem[{Hu et~al.(2016)Hu, Ma, Liu, Hovy, and Xing}]{hu2016harnessing}
Zhiting Hu, Xuezhe Ma, Zhengzhong Liu, Eduard Hovy, and Eric Xing. 2016.
\newblock \href {https://doi.org/10.18653/v1/P16-1228} {Harnessing deep neural
  networks with logic rules}.
\newblock In \emph{Proceedings of the 54th Annual Meeting of the Association
  for Computational Linguistics (Volume 1: Long Papers)}, pages 2410--2420,
  Berlin, Germany. Association for Computational Linguistics.

\bibitem[{Krallinger et~al.(2017)Krallinger, Rabal, Akhondi et~al.}]{chemprot}
Martin Krallinger, Obdulia Rabal, Saber~A Akhondi, et~al. 2017.
\newblock Overview of the biocreative {VI} chemical-protein interaction track.
\newblock In \emph{BioCreative evaluation Workshop}, volume~1, pages 141--146.

\bibitem[{Lee(2013)}]{lee2013pseudo}
Dong-Hyun Lee. 2013.
\newblock Pseudo-label: The simple and efficient semi-supervised learning
  method for deep neural networks.
\newblock In \emph{Workshop on challenges in representation learning, ICML},
  volume~3, page 896.

\bibitem[{Lee et~al.(2020)Lee, Yoon, Kim, Kim, Kim, So, and
  Kang}]{lee2020biobert}
Jinhyuk Lee, Wonjin Yoon, Sungdong Kim, Donghyeon Kim, Sunkyu Kim, Chan~Ho So,
  and Jaewoo Kang. 2020.
\newblock Biobert: a pre-trained biomedical language representation model for
  biomedical text mining.
\newblock \emph{Bioinformatics}, 36(4):1234--1240.

\bibitem[{Lester et~al.(2021)Lester, Al-Rfou, and Constant}]{lester2021power}
Brian Lester, Rami Al-Rfou, and Noah Constant. 2021.
\newblock \href {https://doi.org/10.18653/v1/2021.emnlp-main.243} {The power of
  scale for parameter-efficient prompt tuning}.
\newblock In \emph{Proceedings of the 2021 Conference on Empirical Methods in
  Natural Language Processing}, pages 3045--3059, Online and Punta Cana,
  Dominican Republic. Association for Computational Linguistics.

\bibitem[{Li et~al.(2021{\natexlab{a}})Li, Ding, Shang, McAuley, and
  Feng}]{li2021weakly}
Jiacheng Li, Haibo Ding, Jingbo Shang, Julian McAuley, and Zhe Feng.
  2021{\natexlab{a}}.
\newblock \href {https://doi.org/10.18653/v1/2021.acl-long.352} {Weakly
  supervised named entity tagging with learnable logical rules}.
\newblock In \emph{Proceedings of the 59th Annual Meeting of the Association
  for Computational Linguistics and the 11th International Joint Conference on
  Natural Language Processing (Volume 1: Long Papers)}, pages 4568--4581,
  Online. Association for Computational Linguistics.

\bibitem[{Li and Liang(2021)}]{li2021prefix}
Xiang~Lisa Li and Percy Liang. 2021.
\newblock \href {https://doi.org/10.18653/v1/2021.acl-long.353} {Prefix-tuning:
  Optimizing continuous prompts for generation}.
\newblock In \emph{Proceedings of the 59th Annual Meeting of the Association
  for Computational Linguistics and the 11th International Joint Conference on
  Natural Language Processing (Volume 1: Long Papers)}, pages 4582--4597,
  Online. Association for Computational Linguistics.

\bibitem[{Li et~al.(2021{\natexlab{b}})Li, Shetty, Liu, Zhang, and
  Song}]{li-etal-2021-bertifying}
Yinghao Li, Pranav Shetty, Lucas Liu, Chao Zhang, and Le~Song.
  2021{\natexlab{b}}.
\newblock \href {https://doi.org/10.18653/v1/2021.acl-long.482} {{BERT}ifying
  the hidden {M}arkov model for multi-source weakly supervised named entity
  recognition}.
\newblock In \emph{Proceedings of the 59th Annual Meeting of the Association
  for Computational Linguistics and the 11th International Joint Conference on
  Natural Language Processing (Volume 1: Long Papers)}, pages 6178--6190,
  Online. Association for Computational Linguistics.

\bibitem[{Liang et~al.(2020)Liang, Yu, Jiang, Er, Wang, Zhao, and Zhang}]{bond}
Chen Liang, Yue Yu, Haoming Jiang, Siawpeng Er, Ruijia Wang, Tuo Zhao, and Chao
  Zhang. 2020.
\newblock \href {https://doi.org/10.1145/3394486.3403149} {Bond: Bert-assisted
  open-domain named entity recognition with distant supervision}.
\newblock In \emph{Proceedings of the 26th ACM SIGKDD International Conference
  on Knowledge Discovery \& Data Mining}, page 1054–1064, New York, NY, USA.
  Association for Computing Machinery.

\bibitem[{Lison et~al.(2020)Lison, Barnes, Hubin, and Touileb}]{lison2020named}
Pierre Lison, Jeremy Barnes, Aliaksandr Hubin, and Samia Touileb. 2020.
\newblock \href {https://doi.org/10.18653/v1/2020.acl-main.139} {Named entity
  recognition without labelled data: A weak supervision approach}.
\newblock In \emph{Proceedings of the 58th Annual Meeting of the Association
  for Computational Linguistics}, pages 1518--1533, Online. Association for
  Computational Linguistics.

\bibitem[{Liu et~al.(2021{\natexlab{a}})Liu, Yuan, Fu, Jiang, Hayashi, and
  Neubig}]{liu2021pre}
Pengfei Liu, Weizhe Yuan, Jinlan Fu, Zhengbao Jiang, Hiroaki Hayashi, and
  Graham Neubig. 2021{\natexlab{a}}.
\newblock Pre-train, prompt, and predict: A systematic survey of prompting
  methods in natural language processing.
\newblock \emph{arXiv preprint arXiv:2107.13586}.

\bibitem[{Liu et~al.(2021{\natexlab{b}})Liu, Ji, Fu, Du, Yang, and
  Tang}]{liu2021p}
Xiao Liu, Kaixuan Ji, Yicheng Fu, Zhengxiao Du, Zhilin Yang, and Jie Tang.
  2021{\natexlab{b}}.
\newblock P-tuning v2: Prompt tuning can be comparable to fine-tuning
  universally across scales and tasks.
\newblock \emph{arXiv preprint arXiv:2110.07602}.

\bibitem[{Liu et~al.(2021{\natexlab{c}})Liu, Zheng, Du, Ding, Qian, Yang, and
  Tang}]{liu2021gpt}
Xiao Liu, Yanan Zheng, Zhengxiao Du, Ming Ding, Yujie Qian, Zhilin Yang, and
  Jie Tang. 2021{\natexlab{c}}.
\newblock Gpt understands, too.
\newblock \emph{arXiv preprint arXiv:2103.10385}.

\bibitem[{Liu et~al.(2019)Liu, Ott, Goyal, Du, Joshi, Chen, Levy, Lewis,
  Zettlemoyer, and Stoyanov}]{liu2019roberta}
Yinhan Liu, Myle Ott, Naman Goyal, Jingfei Du, Mandar Joshi, Danqi Chen, Omer
  Levy, Mike Lewis, Luke Zettlemoyer, and Veselin Stoyanov. 2019.
\newblock Roberta: A robustly optimized bert pretraining approach.
\newblock \emph{arXiv preprint arXiv:1907.11692}.

\bibitem[{Loshchilov and Hutter(2019)}]{loshchilov2018adamw}
Ilya Loshchilov and Frank Hutter. 2019.
\newblock \href {https://openreview.net/forum?id=Bkg6RiCqY7} {Decoupled weight
  decay regularization}.
\newblock In \emph{International Conference on Learning Representations}.

\bibitem[{Margatina et~al.(2021)Margatina, Vernikos, Barrault, and
  Aletras}]{margatina2021active}
Katerina Margatina, Giorgos Vernikos, Lo{\"\i}c Barrault, and Nikolaos Aletras.
  2021.
\newblock \href {https://aclanthology.org/2021.emnlp-main.51} {Active learning
  by acquiring contrastive examples}.
\newblock In \emph{Proceedings of the 2021 Conference on Empirical Methods in
  Natural Language Processing}, pages 650--663, Online and Punta Cana,
  Dominican Republic. Association for Computational Linguistics.

\bibitem[{Mekala and Shang(2020)}]{mekala2020contextualized}
Dheeraj Mekala and Jingbo Shang. 2020.
\newblock \href {https://doi.org/10.18653/v1/2020.acl-main.30} {Contextualized
  weak supervision for text classification}.
\newblock In \emph{Proceedings of the 58th Annual Meeting of the Association
  for Computational Linguistics}, pages 323--333, Online. Association for
  Computational Linguistics.

\bibitem[{Meng et~al.(2020)Meng, Zhang, Huang, Xiong, Ji, Zhang, and
  Han}]{meng2020text}
Yu~Meng, Yunyi Zhang, Jiaxin Huang, Chenyan Xiong, Heng Ji, Chao Zhang, and
  Jiawei Han. 2020.
\newblock \href {https://doi.org/10.18653/v1/2020.emnlp-main.724} {Text
  classification using label names only: A language model self-training
  approach}.
\newblock In \emph{Proceedings of the 2020 Conference on Empirical Methods in
  Natural Language Processing (EMNLP)}, pages 9006--9017, Online. Association
  for Computational Linguistics.

\bibitem[{Paszke et~al.(2019)Paszke, Gross, Massa, Lerer, Bradbury, Chanan,
  Killeen, Lin, Gimelshein, Antiga et~al.}]{paszke2019pytorch}
Adam Paszke, Sam Gross, Francisco Massa, Adam Lerer, James Bradbury, Gregory
  Chanan, Trevor Killeen, Zeming Lin, Natalia Gimelshein, Luca Antiga, et~al.
  2019.
\newblock Pytorch: An imperative style, high-performance deep learning library.
\newblock \emph{Advances in neural information processing systems},
  32:8026--8037.

\bibitem[{Perez et~al.(2021)Perez, Kiela, and Cho}]{perez2021true}
Ethan Perez, Douwe Kiela, and Kyunghyun Cho. 2021.
\newblock True few-shot learning with language models.
\newblock \emph{arXiv preprint arXiv:2105.11447}.

\bibitem[{Ratner et~al.(2017)Ratner, Bach, Ehrenberg, Fries, Wu, and
  R{\'e}}]{ratner2017snorkel}
Alexander Ratner, Stephen~H Bach, Henry Ehrenberg, Jason Fries, Sen Wu, and
  Christopher R{\'e}. 2017.
\newblock Snorkel: Rapid training data creation with weak supervision.
\newblock In \emph{Proceedings of the VLDB Endowment. International Conference
  on Very Large Data Bases}, volume~11, page 269. NIH Public Access.

\bibitem[{Safranchik et~al.(2020)Safranchik, Luo, and
  Bach}]{safranchik2020weakly}
Esteban Safranchik, Shiying Luo, and Stephen Bach. 2020.
\newblock Weakly supervised sequence tagging from noisy rules.
\newblock In \emph{Proceedings of the AAAI Conference on Artificial
  Intelligence}, volume~34, pages 5570--5578.

\bibitem[{Schick and
  Sch{\"u}tze(2021{\natexlab{a}})}]{schick-schutze-2021-exploiting}
Timo Schick and Hinrich Sch{\"u}tze. 2021{\natexlab{a}}.
\newblock \href {https://doi.org/10.18653/v1/2021.eacl-main.20} {Exploiting
  cloze-questions for few-shot text classification and natural language
  inference}.
\newblock In \emph{Proceedings of the 16th Conference of the European Chapter
  of the Association for Computational Linguistics: Main Volume}, pages
  255--269, Online. Association for Computational Linguistics.

\bibitem[{Schick and Sch{\"u}tze(2021{\natexlab{b}})}]{schick2021just}
Timo Schick and Hinrich Sch{\"u}tze. 2021{\natexlab{b}}.
\newblock \href {https://doi.org/10.18653/v1/2021.naacl-main.185} {It{'}s not
  just size that matters: Small language models are also few-shot learners}.
\newblock In \emph{Proceedings of the 2021 Conference of the North American
  Chapter of the Association for Computational Linguistics: Human Language
  Technologies}, pages 2339--2352. Association for Computational Linguistics.

\bibitem[{Shen et~al.(2017)Shen, Yun, Lipton, Kronrod, and
  Anandkumar}]{shen-etal-2017-deep}
Yanyao Shen, Hyokun Yun, Zachary Lipton, Yakov Kronrod, and Animashree
  Anandkumar. 2017.
\newblock \href {https://doi.org/10.18653/v1/W17-2630} {Deep active learning
  for named entity recognition}.
\newblock In \emph{Proceedings of the 2nd Workshop on Representation Learning
  for {NLP}}, pages 252--256, Vancouver, Canada. Association for Computational
  Linguistics.

\bibitem[{Van~der Maaten and Hinton(2008)}]{van2008visualizing}
Laurens Van~der Maaten and Geoffrey Hinton. 2008.
\newblock Visualizing data using t-sne.
\newblock \emph{Journal of machine learning research}, 9(11).

\bibitem[{Varma and R{\'e}(2018)}]{varma2018snuba}
Paroma Varma and Christopher R{\'e}. 2018.
\newblock Snuba: Automating weak supervision to label training data.
\newblock In \emph{Proceedings of the VLDB Endowment. International Conference
  on Very Large Data Bases}, volume~12, page 223. NIH Public Access.

\bibitem[{Xie et~al.(2016)Xie, Girshick, and Farhadi}]{xie2016unsupervised}
Junyuan Xie, Ross Girshick, and Ali Farhadi. 2016.
\newblock Unsupervised deep embedding for clustering analysis.
\newblock In \emph{International conference on machine learning}, pages
  478--487. PMLR.

\bibitem[{Yu et~al.(2021{\natexlab{a}})Yu, Kong, Zhang, Zhang, and
  Zhang}]{yu2021atm}
Yue Yu, Lingkai Kong, Jieyu Zhang, Rongzhi Zhang, and Chao Zhang.
  2021{\natexlab{a}}.
\newblock Atm: An uncertainty-aware active self-training framework for
  label-efficient text classification.
\newblock \emph{arXiv preprint arXiv:2112.08787}.

\bibitem[{Yu et~al.(2021{\natexlab{b}})Yu, Zuo, Jiang, Ren, Zhao, and
  Zhang}]{yu2020fine}
Yue Yu, Simiao Zuo, Haoming Jiang, Wendi Ren, Tuo Zhao, and Chao Zhang.
  2021{\natexlab{b}}.
\newblock \href {https://doi.org/10.18653/v1/2021.naacl-main.84} {Fine-tuning
  pre-trained language model with weak supervision: A contrastive-regularized
  self-training approach}.
\newblock In \emph{Proceedings of the 2021 Conference of the North American
  Chapter of the Association for Computational Linguistics: Human Language
  Technologies}, pages 1063--1077, Online. Association for Computational
  Linguistics.

\bibitem[{Zhang et~al.(2022)Zhang, Hsieh, Yu, Zhang, and Ratner}]{zhang2022}
Jieyu Zhang, Cheng-Yu Hsieh, Yue Yu, Chao Zhang, and Alexander Ratner. 2022.
\newblock A survey on programmatic weak supervision.
\newblock \emph{arXiv preprint arXiv:2202.05433}.

\bibitem[{Zhang et~al.(2021)Zhang, Yu, Li, Wang, Yang, Yang, and
  Ratner}]{zhang2021wrench}
Jieyu Zhang, Yue Yu, Yinghao Li, Yujing Wang, Yaming Yang, Mao Yang, and
  Alexander Ratner. 2021.
\newblock \href {https://openreview.net/forum?id=Q9SKS5k8io} {{WRENCH}: A
  comprehensive benchmark for weak supervision}.
\newblock In \emph{Thirty-fifth Conference on Neural Information Processing
  Systems Datasets and Benchmarks Track}.

\bibitem[{Zhang et~al.(2020)Zhang, Yu, and Zhang}]{zhang2020seqmix}
Rongzhi Zhang, Yue Yu, and Chao Zhang. 2020.
\newblock \href {https://doi.org/10.18653/v1/2020.emnlp-main.691} {{S}eq{M}ix:
  Augmenting active sequence labeling via sequence mixup}.
\newblock In \emph{Proceedings of the 2020 Conference on Empirical Methods in
  Natural Language Processing (EMNLP)}, pages 8566--8579, Online. Association
  for Computational Linguistics.

\bibitem[{Zhang et~al.(2015)Zhang, Zhao, and LeCun}]{zhang2015character}
Xiang Zhang, Junbo Zhao, and Yann LeCun. 2015.
\newblock Character-level convolutional networks for text classification.
\newblock \emph{Advances in neural information processing systems},
  28:649--657.

\bibitem[{Zhang et~al.(2017)Zhang, Zhong, Chen, Angeli, and
  Manning}]{zhang2017position}
Yuhao Zhang, Victor Zhong, Danqi Chen, Gabor Angeli, and Christopher~D.
  Manning. 2017.
\newblock \href {https://doi.org/10.18653/v1/D17-1004} {Position-aware
  attention and supervised data improve slot filling}.
\newblock In \emph{Proceedings of the 2017 Conference on Empirical Methods in
  Natural Language Processing}, pages 35--45, Copenhagen, Denmark. Association
  for Computational Linguistics.

\bibitem[{Zhao et~al.(2021)Zhao, Ding, and Feng}]{glara}
Xinyan Zhao, Haibo Ding, and Zhe Feng. 2021.
\newblock \href {https://aclanthology.org/2021.eacl-main.318} {{GL}a{RA}:
  Graph-based labeling rule augmentation for weakly supervised named entity
  recognition}.
\newblock In \emph{Proceedings of the 16th Conference of the European Chapter
  of the Association for Computational Linguistics: Main Volume}, pages
  3636--3649, Online. Association for Computational Linguistics.

\bibitem[{Zhou et~al.(2020)Zhou, Lin, Lin, Wang, Du, Neves, and
  Ren}]{zhou2020nero}
Wenxuan Zhou, Hongtao Lin, Bill~Yuchen Lin, Ziqi Wang, Junyi Du, Leonardo
  Neves, and Xiang Ren. 2020.
\newblock Nero: A neural rule grounding framework for label-efficient relation
  extraction.
\newblock In \emph{The Web Conference}, pages 2166--2176.

\end{thebibliography}
\bibliographystyle{acl_natbib}

\appendix

\clearpage
\newpage

\begin{table*}[]
    \centering
    \scalebox{0.8}{
\begin{tabular}{lc}\hline
     \textbf{Rule}& \textbf{Label}  \\\midrule
     If \texttt{[Mask]} prediction is in \{\textit{Economic, Deal, Business, Market}\} & Business \\
     If \texttt{[Mask]} prediction is in \{\textit{Microsoft, Tech, Software}\}& Sci/Tech\\
    If \texttt{[Mask]} prediction is in \{\textit{African, Global, World}\} & World\\
    If \texttt{[Mask]} prediction is in \{\textit{NFL, Sports, Team, Football}\} & Sports\\\midrule
    If entity pair $==$ (Organization, Organization) and \texttt{[Mask]} prediction is in \{\textit{formerly, called, aka}\} & org:alternate\_names\\
    If entity pair $==$ (Person, Organization) and \texttt{[Mask]} prediction is in \{\textit{founded, established, started}\} & org:founded\_by\\
    If entity pair $==$ (Person, Title) and \texttt{[Mask]} prediction is in \{\textit{president, head, chairman, director}\} & org:top\_members\\
    If entity pair $==$ (Person, City) and \texttt{[Mask]} prediction is in \{\textit{moved to, lived in, grew in}\} & per:city\_of\_residence\\\bottomrule
\end{tabular}
}
    \caption{\small{More rule examples on the text classification dataset AG News and the relation extraction dataset TACRED.}}
    \label{tab:more_example_rules}
\end{table*}

\section{Dataset Details}
\begin{table*}[!htb]
  \centering
  \scalebox{0.95}{
    \begin{tabular}{cccccc}
    \toprule
        \bf Dataset & \bf  Task & \bf  Domain & \bf  \# Class  &\bf   \# Train & \bf \# Test  \\\midrule
        TACRED & Relation Extraction & Web Text &41 & 68,124 & 15,509\\
        DBPedia & Ontology Classification& Wikipedia Text & 14 & 560,000 & 70,000\\
        Chemprot & Chemical-protein Interaction Prediction& Biology &  10 & 5,400 & 1,400\\
        AG News & News Topic Classification& News & 4 & 120,000 & 7,600\\
        \bottomrule
    \end{tabular}
}
  \caption{Dataset statistics.}
  \label{tab:dataset}
\end{table*}

\noindent\textbf{Weak sources}
\label{appendix_weak_sources}
For each dataset above, we have an existing weak source that uses labeling rules to generate weakly labeled data.
\begin{enumerate}[leftmargin=*]
    \item \textit{TACRED}: We use the rules in \citet{zhou2020nero} for the relation extraction task. Their rules are in the form of relation phrases, which include the entity pair and a keyword.
    \item \textit{DBPedia}: We use the keywords provided in~\cite{meng2020text} as the labeling rules. Such keywords are indicative to the categories, where the words for the same category have close semantics. 
     \item \textit{AGNews, ChemProt}: We use the rules in~\citet{zhang2021wrench} as the labeling fucntions. They also extract lexical patterns for weak supervision.
\end{enumerate}

\section{Hyper-parameters}
We show the hyper-parameter configuration in Table~\ref{tab:hyperparameter}. We search the batch size in $\{8, 16, 32, 64, 128\}$, AND the coefficient $\alpha$ between $[0, 1]$ with an interval of $0.25$. For the optimizer, we use AdamW \cite{loshchilov2018adamw} and choose learning rate from $\{5 \times 10^{-6}, 1\times 10^{-5}, 2\times 10^{-5}\}$. We keep the number of iterations as 10 for all the tasks and show the top-$10$ candidate rules to solicit human feedback. ChemProt is a special case where we present the top-$20$ candidate rules, because this task is more domain-specific than the others, and the involved human annotators have no relevant domain background.
\begin{table*}[!htb]
	\begin{center}
	\scalebox{1}{
		\begin{tabular}{c|c|c|c|c}
			\toprule 
			\bf Hyper-parameter &\bf TACRED & \bf DBpedia & \bf ChemProt  & \bf AG News \\\hline
			 Maximum Tokens  & 128 & 256 & 512 & 128 \\ \hline 
			 Batch Size  & 32 & 32 & 8 & 32\\ \hline
			 Learning Rate & $2\times10^{-5}$ & $10^{-5}$ & $10^{-5}$ & $10^{-5}$\\\hline
			 Dropout Rate & 0.2 & 0.1& 0.1&0.1\\\hline
			 \# Iterations &10&10&10&10\\\hline
			  $\alpha$ &0.5& 0.25& 0.5& 0.25 \\\hline
			 $k$ &10&10&20&10 \\ \bottomrule
		\end{tabular}}
	\end{center}
	\caption{Hyper-parameter configurations.}
	\label{tab:hyperparameter}
\end{table*}

\section{Implementation Setting}
\label{appendix_para_setting}
We test our code on the System \textit{Ubuntu 18.04.4 LTS} with CPU: \textit{Intel(R) Xeon(R) Silver 4214 CPU
@ 2.20GHz} and \textit{GPU: NVIDIA GeForce RTX 2080}.
We implement our method using Python 3.6 and PyTorch 1.2~\citep{paszke2019pytorch}.

\section{Interactive baselines}
\label{iter_baseline}
For interactive learning, We include an interactive weak supervision framework IWS \cite{boecking2020interactive}, the most recent AL method CAL \cite{margatina2021active} and the entropy-based AL as baselines. Our goal is 1) to compare the annotation cost of rule-level annotation and instance-level annotation; 2) to compare the model performance with the same annotation budget. 

Because IWS is designed for the binary classification problem, we revise its implementation by integrating multiple binary predictions for multi-class tasks. Specifically, we obtain the predicted probability over all categories from each classifier, and select the category with the highest probability as the final prediction. When the number of category is large, this approach becomes cumbersome as training multiple classifiers is time-consuming. Therefore, we only run IWS on AG News, which has 4 categories. We report the results of these interactive methods in Section~\ref{sec_user_exp} and the following Appendix~\ref{user_study}.

\section{User Study}
\label{user_study}
\begin{figure}[!htb]
    \centering
    \includegraphics[scale=0.4]{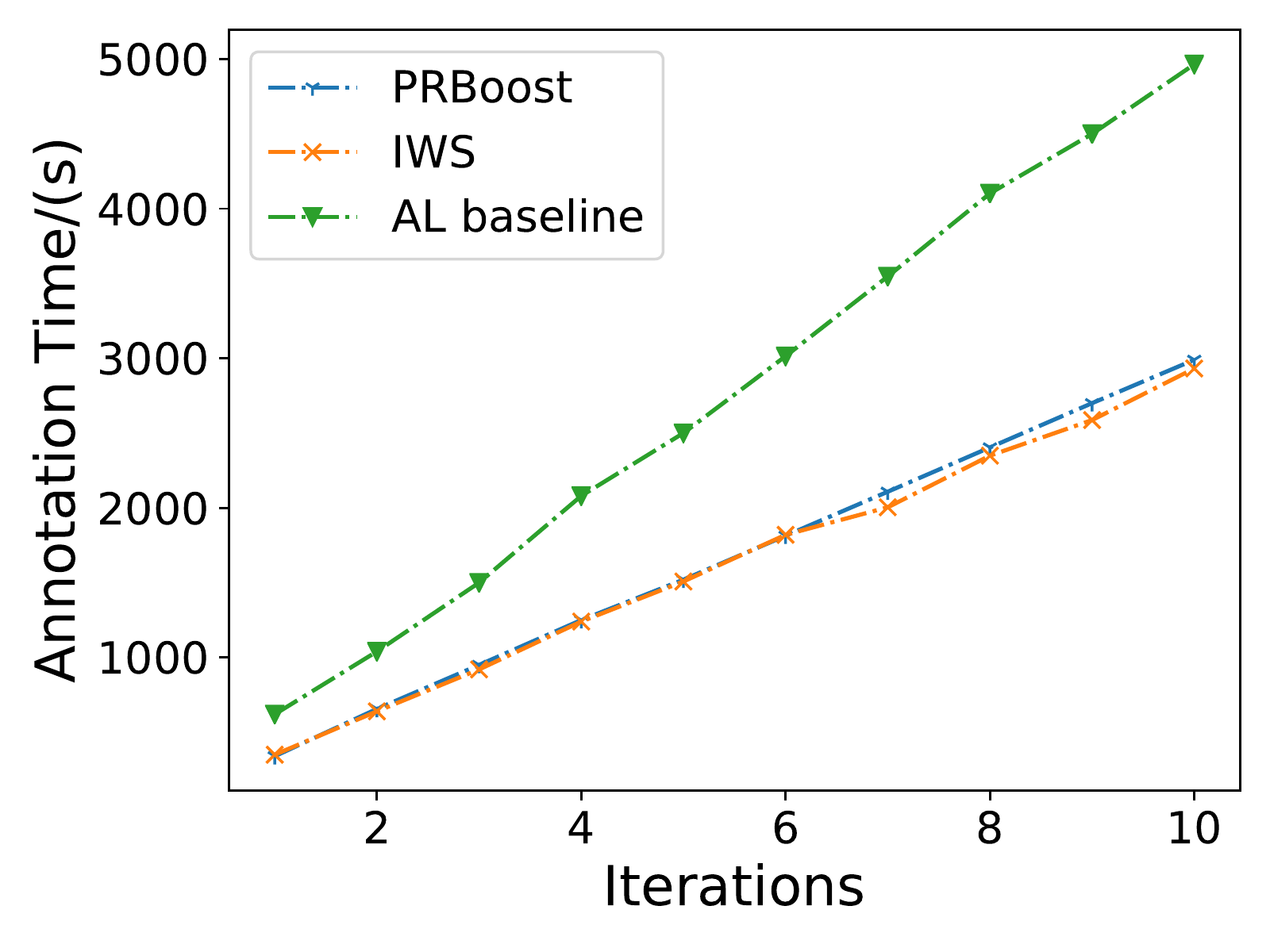}
    \caption{Annotation cost of interactive methods measured by annotation time on AGNews. Both \ours and IWS use rule-level annotation, while AL baselines use instance-level annotation.}
    \label{fig:eval_time}
\end{figure}
In this user study, we aim to measure the annotation cost and the inter-annotator agreement during the rule annotation process. 
We ask three human annotators to participate in the 10-iteration experiment. In each iteration, humans are asked to annotate 100 candidate rules.
We count the time in each iteration and their binary decisions on each candidate rule. 
The averaged annotation time is compared in Section~\ref{sec_user_exp} and we present more details in Figure~\ref{fig:eval_time}.

The rule-level annotation agreement is measured by the Fleiss' kappa $\kappa$ defined as
\begin{equation}
 \setlength{\abovedisplayskip}{2pt}
\setlength{\belowdisplayskip}{2pt}
    \kappa = (\bar P-\bar P_e)/(1-\bar P_e),
\end{equation}
where $\bar P$ measures the annotation agreement over all categories, and $\bar P_e$ computes the quadratic sum of
the proportion of assignments to each category. The results in Section\ref{sec_user_exp} demonstrate that human annotators can achieve substantial agreement on rule-level annotation.

The rules to be annotated are generated from open-source PLMs and public data. We believe this rule-level annotation process will not amplify any bias in the original data. We do not foresee any ethical issues or direct social consequences.

\section{Model Ensemble}
In practice, we keep $\alpha_t$ for each weak model as same during the model ensemble. 
Equation~\ref{eq:ensemble} weights each weak model $m_t$ by a computed coefficient $\alpha_t$. Intuitively, the weak model $m_t$ with higher $\alpha_t$ impacts the ensemble results more. This paradigm is proved to be effective under fully-supervised settings, but we found it is not directly applicable in WSL. Since we initialize a model $m_0$ on the given weak source and it can achieve a relatively strong performance (much better than random guess), \textit{i.e.}, the error rate ${err}_0$ is low. It makes a high $\alpha_0$ based on Equation~\ref{eq:alpha}, so the initialized model will dominate the following prediction, thus limiting the effectiveness of the model ensemble. Therefore, we assign the same weight to each weak model but still follow the design of identifying large-error instances. This is reasonable as the weight $w_i$ computed by Equation~\ref{eq_weights} still reflects the model weakness and can guide the rule proposal. By discovering rules based on the large-error instances, we iteratively complement the feature regimes through the model training on rule-matched data and strengthen the ensemble model.

\end{document}